\documentclass{article}

\usepackage{microtype}
\usepackage{graphicx}
\usepackage{booktabs} % for professional tables

\usepackage{hyperref}

\usepackage[accepted]{icml2025}

\usepackage{amsmath}
\DeclareMathOperator*{\argmin}{arg\,min}
\usepackage{amssymb}
\usepackage{mathtools}
\usepackage{amsthm}

\usepackage[capitalize,noabbrev]{cleveref}

\usepackage{multirow}
\usepackage{xspace}
\usepackage{enumitem}
\usepackage{algorithmic}
\usepackage{algorithm}
\usepackage{url}
\usepackage{color}
\usepackage{wrapfig}
\usepackage{stmaryrd}
\usepackage{subcaption}
\usepackage{amsmath}

\theoremstyle{plain}
\newtheorem{theorem}{Theorem}[section]

\theoremstyle{definition}

\theoremstyle{remark}

\usepackage[textsize=tiny]{todonotes}

\newcommand{\proj}{SafeDiffCon\xspace}

\renewcommand{\t}{k}
\newcommand{\T}{K}
\renewcommand{\u}{\mathbf{u}}

\newcommand{\w}{\mathbf{w}}

\newcommand{\x}{\mathbf{x}}

\newcommand{\bepsilon}{\bm{\epsilon}} % noise

\def\J{\mathcal{J}} % Objective
\def\W{\mathcal{W}} % J+safe
\def\G{\mathcal{G}} % Guidance
\def\S{\mathcal{S}} % Score Set
\def\score{s}
\def\pu{\mathbf{u}_\theta} % predicted u
\def\R{\mathcal{R}} % Unsafe Rate

\usepackage{multirow}
\usepackage{pifont}
\usepackage{array}
\usepackage{bm}
\icmltitlerunning{From Uncertain to Safe: Conformal Adaptation of Diffusion Models for Safe PDE Control}

\begin{document}

\twocolumn[
% \icmltitle{Uncertain to Safe: Conformal Adaptaion of Diffusion Models for Safe PDE Control}
\icmltitle{\makebox[\textwidth][c]{From Uncertain to Safe: Conformal Adaptation of Diffusion Models} \\ \makebox[\textwidth][c]{for Safe PDE Control}}



\icmlsetsymbol{equal}{*}

\begin{icmlauthorlist}
\icmlauthor{Peiyan Hu}{equal,intern,cas}
\icmlauthor{Xiaowei Qian}{equal,intern,westlake}
\icmlauthor{Wenhao Deng}{westlake}
\icmlauthor{Rui Wang}{intern,fudan}
\icmlauthor{Haodong Feng}{westlake}
\icmlauthor{Ruiqi Feng}{westlake}
\icmlauthor{Tao Zhang}{westlake}
\icmlauthor{Long Wei}{westlake}
\icmlauthor{Yue Wang}{zgc}
\icmlauthor{Zhi-Ming Ma}{cas}
\icmlauthor{Tailin Wu}{westlake}
\end{icmlauthorlist}

\icmlaffiliation{intern}{Work done as an intern at Westlake University}
\icmlaffiliation{westlake}{School of Engineering, Westlake University, Hangzhou, China}
\icmlaffiliation{cas}{Academy of Mathematics and Systems Science, Chinese Academy of Sciences, Beijing, China}
\icmlaffiliation{fudan}{Fudan University, Shanghai, China}
\icmlaffiliation{zgc}{Zhongguancun Academy, Beijing, China}

\icmlcorrespondingauthor{Tailin Wu}{wutailin@westlake.edu.cn}

\icmlkeywords{Machine Learning, ICML}

\vskip 0.3in
]



\printAffiliationsAndNotice{\icmlEqualContribution} % otherwise use the standard text.

\begin{abstract}

The application of deep learning for partial differential equation (PDE)-constrained control is gaining increasing attention. However, existing methods rarely consider safety requirements crucial in real-world applications. To address this limitation, we propose \emph{\underline{Safe} \underline{Diff}usion Models for PDE \underline{Con}trol} (\proj), which introduce the uncertainty quantile as model uncertainty quantification to achieve optimal control under safety constraints through both post-training and inference phases. Firstly, our approach post-trains a pre-trained diffusion model to generate control sequences that better satisfy safety constraints while achieving improved control objectives via a reweighted diffusion loss, which incorporates the uncertainty quantile estimated using conformal prediction. Secondly, during inference, the diffusion model dynamically adjusts both its generation process and parameters through iterative guidance and fine-tuning, conditioned on control targets while simultaneously integrating the estimated uncertainty quantile.
We evaluate \proj on three control tasks: 1D Burgers' equation, 2D incompressible fluid, and controlled nuclear fusion problem. Results demonstrate that \proj is the only method that satisfies all safety constraints, whereas other classical and deep learning baselines fail. Furthermore, while adhering to safety constraints, \proj achieves the best control performance. The code can be found at \url{https://github.com/AI4Science-WestlakeU/safediffcon}.
 
\end{abstract}

\section{Introduction}

The control of physical systems described by partial differential equations (PDEs) is crucial in various scientific and engineering domains, including fluid dynamics \citep{hinze2001second}, controlled nuclear fusion \citep{schuster2006role}, and mathematical finance \citep{soner2004stochastic}. While traditional control algorithms have been extensively studied over the years \citep{1580152, protas2008adjoint}, advancements in neural networks have led to the development of numerous deep learning-based approaches \citep{farahmand2017deep, holl2020learning, hwang2022solving}. Among these, diffusion models have gained increasing attention to PDE control due to their ability to model high-dimensional and nonlinear data \citep{vahdat2022lion, li2024synthetic}, and have demonstrated impressive performance in various control tasks \citep{chi2023diffusion, wei2024generative, hu2024wavelet}.

However, existing deep learning-based methods for PDE control generally overlook \textit{safety} -- meaning ensuring control sequences satisfy predefined constraints, mitigating risks, and preventing hazards \citep{dawson2022safe, liu2023datasets}. In real-world applications, controlling PDE systems particularly requires addressing safety concerns \citep{735940, argomedo2013lyapunov}. For instance, in fluid dynamics, minor errors can cause turbulence or structural damage, while in controlled nuclear fusion, violating safety constraints may lead to catastrophic failures. This lack of safety enforcement remains a major bottleneck in applying deep learning to scientific and engineering problems, posing a critical challenge for high-stakes applications.

Despite its importance, safe PDE control remains challenging. First, ensuring safety requires preventing methods without formal guarantees from interacting with the environment, restricting the approach to an offline setting with pre-collected data. However, such data is often suboptimal and includes unsafe samples, creating a significant gap between the observed distribution and the near-optimal, safe distribution \citep{xu2022constraints, liu2023datasets}. 
Second, the method must resolve the inherent conflict between optimizing control performance and satisfying safety constraints (\emph{e.g.}, aggressive control policies that reduce error may violate safety thresholds) \citep{liu2023datasets, zheng2024safe}.

To address these challenges, we propose \emph{Safe Diffusion Models for PDE Control} (\proj), a method that adapts diffusion models to varying safety constraints by quantifying their uncertainty. 
In offline settings, training data are often sub-optimal and unsafe, which causes the distribution learned by diffusion models to deviate significantly from the desired optimal and safe distribution. Inspired by conformal prediction \citep{vovk2005algorithmic, Tibshirani2019ConformalPU}, we propose quantifying diffusion models' safety uncertainty under distribution shifts and incorporating it through both post-training and inference phases.
%using a portion of split-out training data, referred to as the \textit{calibration set}. 
This uncertainty quantification, termed as \textit{uncertainty quantile}, allows us to compute a \emph{conformal interval}, within which the safety score of the diffusion model interacting with the real environment is expected to lie. By leveraging this uncertainty quantification for predicted safety, diffusion models can be optimized with a clear safety optimization objective, to satisfy various safety constraints by constraining their conformal intervals within safe boundaries. 
Specifically, we incorporate uncertainty quantification into both the post-training and inference-time fine-tuning processes. Technically, we perform \textit{post-training with reweighted loss}, where the weighting accounts for the uncertainty quantile of safety instead of the original predicted safety score. This refinement improves the model's control performance while enhancing its adherence to safety constraints, leading to the generation of more optimal and safer control sequences. To further ensure the safety of PDE control, we introduce \textit{inference-time fine-tuning}, where the diffusion model is iteratively adjusted with guidance and fine-tuning based on specific control targets and uncertainty quantile to improve the safety of the final outputs within few iterations.

Our main contributions are as follows:
\textbf{(1)} We introduce safety constraints into deep learning-based control of PDE systems and propose our method \proj.
\textbf{(2)} To address significant distribution shifts, we quantify the safety uncertainty of diffusion models and design a conformal adaptation process, which incorporates post-training with a reweighted loss and inference-time fine-tuning to promote a safer and more optimal output distribution. 
\textbf{(3)} We design safe control tasks in the contexts of the 1D Burgers' equation, 2D incompressible fluid, and the controlled nuclear fusion scenario, and evaluate various methods. Extensive experiments on these datasets demonstrate that \proj is the only method to successfully meet all safety constraints while achieving superior control objectives.

\section{Related Work}
\subsection{Control of PDE Systems}
The development of control methods in physical systems is critical across various scientific and engineering areas, including PID \citep{1580152}, supervised learning (SL) \citep{holl2020learning,hwang2022solving}, reinforcement learning (RL) \citep{farahmand2017deep, pan2018reinforcement, rabault2019artificial}, and physics-informed neural networks (PINNs) \citep{mowlavi2023optimal}. Among these, PID is one of the earliest and most widely used method \citep{johnson2005pid}, known for its simplicity and effectiveness in regulating physical systems; however, it faces challenges in parameter tuning and struggles with highly nonlinear or time-varying systems. Model Predictive Control (MPC) \cite{garcia1989model,findeisen2002introduction,schwenzer2021review} is another well-known control strategy that optimizes a control sequence over a finite time horizon by solving an optimization problem at each time step, using a dynamic model of the system. As it requires solving an optimization problem in real-time at each time step, it may have a high computational cost and struggle to handle high-dimensional, complex systems effectively.
% With the advancement of deep learning, SL \citep{holl2020learning} has been applied to optimize control sequences through backpropagation over entire trajectories, but it lacks the adaptability to dynamic environments since it is typically trained on fixed datasets. 
% To overcome the above issue, RL enhances adaptability by leveraging diverse datasets or interactions with the environment, achieving notable success in controlling physical systems, including fluid dynamics \citep{novati2017synchronisation, feng2023control}, underwater devices \citep{zhang2022pde, feng2024efficient}, and nuclear fusion \citep{degrave2022magnetic}. 
Furthermore, the adjoint method \citep{protas2008adjoint} and PINNs \citep{mowlavi2023optimal} are also incorporated in PDE control, but they require an explicit form of the PDE. Currently, the diffusion model used in physical systems' control \citep{wei2024generative, hu2024wavelet} integrates the learning of entire state trajectories and control sequences, enabling global optimization that incorporates the physical information learned by the model. However, it does not consider the important cases where safety constraints are required. 

In the field of reinforcement learning (RL), several safe offline RL algorithms have emerged in recent years. CPQ \citep{xu2022constraints} is the first practical safe offline RL method that assigns high costs to out-of-distribution (OOD) and unsafe actions and updates the value function as well as the policy only with safe actions. COptiDICE \citep{lee2022coptidice} is a DICE-based method and corrects the stationary distribution. CDT \citep{liu2023constrained} takes the decision transformer to solve safe offline RL problems as multi-objective optimization. More recent methods such as TREBI \citep{lin2023safe} and FISOR \citep{zheng2024safe} address this problem through diffusion model planning, leveraging the capability of diffusion models to model high-dimensional data. However, none of these algorithms take into account the uncertainty of the predicted safety from the perspective of conformal prediction, nor do they address PDE-constrained scenarios.

% \subsection{Safe Offline Reinforcement Learning}
% Recently, the offline setting has attracted attention in the field of safe reinforcement learning (RL), as it avoids generating dangerous behaviors through direct interaction with the environment \citep{achiam2017constrained, zhang2020first, stooke2020responsive, liu2022constrained}. CPQ \citep{xu2022constraints} is the first practical safe offline RL method that assigns high costs to OOD and unsafe actions and updates the value function as well as the policy only with safe actions. COptiDICE \citep{lee2022coptidice} is a DICE-based method and corrects the stationary distribution. CDT \citep{liu2023constrained} takes the decision transformer to solve safe offline RL problems as multi-objective optimization. However, these methods lack the ability to model high-dimensional state space. More recent methods like TREBI \citep{lin2023safe} and FISOR \citep{zheng2024safe} solve it through diffusion model planning. But they do not consider the upper bound of the safety score in a probabilistic sense and differ significantly from \proj in terms of the algorithm.

\subsection{Conformal Prediction}

% Conformal prediction \citep{vovk2005algorithmic} is a statistical framework that constructs prediction intervals guaranteed to contain the true label with a specified probability. However, its core assumption of exchangeability is often violated in real-world scenarios \citep{chernozhukov2018exact, hendrycks2018using} due to distribution shifts, compromising its validity. To address this issue, recent studies \citep{Cauchois2024Robust} have extended conformal prediction to accommodate various types of distribution shifts. For instance, \citet{Tibshirani2019ConformalPU} proposes weighted conformal prediction to handle covariate shift, where the training and test data distributions differ. In addition, \citet{Podkopaev2021Disfree} introduces reweighted conformal prediction and calibration techniques to address label shift using unlabeled target data. Moreover, adaptive conformal inference \citep{gibbs2021adaptive} provides valid prediction sets in online settings with unknown, time-varying distribution shifts without relying on exchangeability. Inspired by these approaches, our \proj effectively handles distribution shifts between pre-collected data and the target distribution by partitioning the training dataset and computing a weighted scoring set.

Conformal prediction \citep{vovk2005algorithmic} is a statistical framework that constructs prediction intervals guaranteed to contain the true label with a specified probability. Its validity could be compromised, however, by the violation of the core assumption of exchangeability due to distribution shifts in real-world scenarios \citep{chernozhukov2018exact, hendrycks2018using}. 
Recent studies \citep{Cauchois2024Robust} have extended conformal prediction to accommodate various distribution shifts. For example, \citet{Tibshirani2019ConformalPU} proposed weighted conformal prediction to handle covariate shift, where training and test data distributions differ. \citet{Podkopaev2021Disfree} introduced reweighted conformal prediction and calibration techniques to address label shift using unlabeled target data. Adaptive conformal inference \citep{gibbs2021adaptive} provides valid prediction sets in online settings with unknown, time-varying distribution shifts without relying on exchangeability. 
Inspired by conformal prediction, we measure uncertainty based on the distribution of data generated by diffusion models for control problems. This is also the first time the perspective of conformal prediction has been applied to address safety control problems.
% Inspired by previous approaches, \proj establishes an upper bound on the confidence level by maintaining a weighted score set. Our method ensures the true safety value for a given control sequence lies within the bound without requiring additional assumptions about the model or data distribution, effectively addressing distribution shifts between pre-collected data and the target distribution.

\section{Preliminary}
\begin{figure*}[t]
% \vspace{-30pt}  % this might break margins, but worth trying.
\begin{center}
    \includegraphics[scale=0.75]{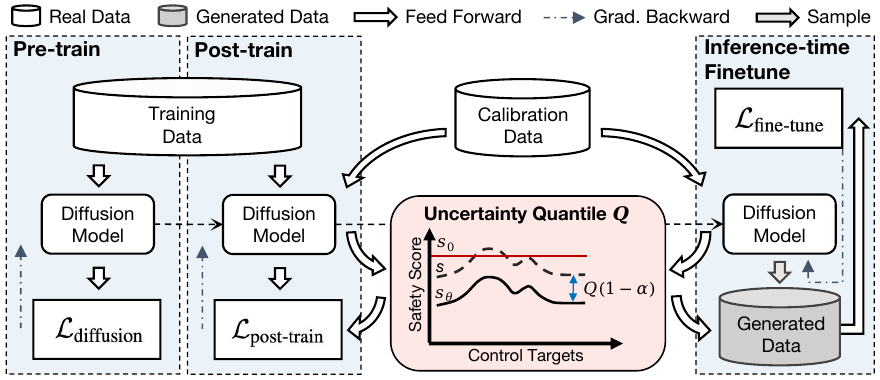}
\end{center}
\vspace{-10pt}
\caption{\textbf{Overview of \proj}. First, we pre-train a diffusion model $p_{\theta}$ on the training data. Then, combined with the uncertainty quantile, we post-train the model to steer its distribution to safer regions with better control objectives. Finally, to improve performance and safety for specific control tasks, we conduct inference-time fine-tuning, again incorporating the uncertainty quantile into the process.}
\vspace{-10pt}
\label{fig:overview}
\end{figure*}

\subsection{Problem Setup}
We consider the following safe control problem of PDE-constrained systems:
\begin{equation}
    \w^*=\argmin_\w\J(\u,\w)\quad\text{s.t.}\quad\mathcal{C}(\u,\w)=0,\quad\score(\u)\leq \score_0,
    \label{eq:origion_optimization}
\end{equation}
where $\u(t,\x):[0,T]\times \Omega\mapsto \mathbb{R}^{d_\u}$ is the system's state trajectory with dimension ${d_\u}$ and $\w(t,\x):[0,T]\times \Omega\mapsto \mathbb{R}^{d_\w}$ is the external control signal with dimension ${d_\w}$. They are both defined on the time range $[0,T]\subset\mathbb{R}$ and spatial domain $\Omega\subset \mathbb{R}^{D}$. 
$\J(\u,\w)$ is the objective of the control problem, and $\mathcal{C}(\u,\w)=0$ is the PDE constraint.

As for the safety constraint, $\score(\u)$ is the safety score and $\score_0$ is the bound of the safety score. 
%Although $\score$ is generally a direct function of $\u$, since we ultimately need the model's output $\w$, and under physical constraints, $\u$ is a function of $\w$, we will therefore express it as $\score(\w)$ in the following paper. 
We need to minimize the control objective while satisfying PDE constraints and constraining the safety score to stay below the bound, which requires a careful balance between safety and control performance. Furthermore, it is important to note that safety and control performance are not on equal footing, and the pursuit of a better objective should be built upon ensuring safety \cite{knight2002safety, cheng2019end}.

\subsection{Diffusion Models and Diffusion Control}
\label{sec:ddpm}

Diffusion models \citep{ho2020denoising} learn data distribution from data in a generative way. They present impressive performance in a broad range of generation tasks. Diffusion models involve diffusion/denoising processes: the diffusion process $q(\x^{\t+1}|\x^\t)=\mathcal{N}(\x^{\t+1};\sqrt{\alpha_\t}\x_\t,(1-\alpha_\t)\mathbf{I})$
corrupts the data distribution $p(\x_0)$ to a prior distribution $\mathcal{N}(\mathbf{0}, \mathbf{I})$, and the denoising process $p_{\theta}(\x^{\t-1}|\x^\t)=\mathcal{N}(\x^{\t-1};\mu_\theta(\x^\t,\t),\sigma_\t \mathbf{I})$ makes sampling in a reverse direction. Here $\t$ is the diffusion/denoising step, $\{\alpha_\t\}_{\t=1}^\T$ and $\{\sigma_\t\}_{\t=1}^\T$ are the noise and variance schedules. In practice, a denoising network $\bepsilon_{\theta}$ is trained to estimate the noise to be removed in each step. 
During inference, the iterative removal of $\bepsilon_{\theta}$ from the prior distribution could generate a new sample that follows the data distribution $p(\x)$. 

Recently, diffusion models \citep{wei2024generative, hu2024wavelet} are applied to solve the control problem as in Eq. \ref{eq:origion_optimization} without the safety constraint $\score(\u)\leq \score_0$. For brevity, we only summarize the light version. These methods transform the physical constraint to a parameterized energy-based model (EBM) $E_{\theta}(\u,\w)$ with the correspondence  $p(\u,\w)\propto \exp({-E_{\theta}(\u,\w)})$. Then the problem is converted to an unconstrained optimization over $\u$ and $\w$ for all physical time steps simultaneously:
\vspace{-2pt}
\begin{gather}
\label{eq:joint_optimization}
\u^*, \w^* = \argmin_{\u, \w}\left[E_\theta(\u,\w) + \lambda\cdot \mathcal{J}(\u,\w)\right],
\end{gather}
where $\lambda$ is a hyperparameter. 
To optimize $E_{\theta}$, a denoising network $\bepsilon_\theta$ is trained to approximate $\nabla_{\u,\w} E_\theta(\u,\w)$ by the following loss:
\begin{align}
\label{eq:training_obj}
\mathcal{L}_\textrm{diffusion}&=\mathbb{E}_{\t\sim U(1,\T),(\u,\w)\sim p(\u,\w),\bepsilon\sim \mathcal{N}(\mathbf{0},\mathbf{I})} \notag \\ 
&\quad [\|\mathbf{\bepsilon} - \bepsilon_\theta(\sqrt{\bar{\alpha}_\t}[\u,\w] + \sqrt{1-\bar{\alpha}_\t} \bepsilon,\t)\|_2^2],
\end{align}
where $\Bar{\alpha}_\t:=\prod_{i=1}^\t\alpha_i$. After $\bepsilon_\theta$ is trained, Eq. \ref{eq:joint_optimization} can be optimized by sampling from an initial sample $(\u^\T,\w^\T)\sim\mathcal{N}(\mathbf{0},\mathbf{I})$, and iteratively running the following process\footnote{In this paper, we use $\mathbf{u}$ to represent state trajectory across time, and $\mathbf{w}$ to represent control signal sequence across time. $\mathbf{u}^k$ denotes the full system trajectory across time at denoising step $k$.}
\begin{align}
\label{eq:1ddpm_inference}
 (\u^{\t-1},\w^{\t-1}) = (\u^{\t},\w^{\t}) - \eta_k\cdot\bepsilon_\theta([\u^{\t},\w^{\t}],\t) \notag \\ 
 +\lambda_k\nabla_{\u,\w}\mathcal{G}(\hat{\u}^\t,\hat{\w}^\t) + \xi, \quad \mathbf{\xi} \sim \mathcal{N} \bigl(\mathbf{0}, \sigma^2_\t \mathbf{I} \bigl)
\end{align}
under the guidance of $\G=\J$ for $\t=\T,\T-1,..., 1$. Here 
% $\sigma^2_t$ and $\eta$ correspond to noise schedules and scaling factors, respectively, and 
$[\hat{\u}^\t,\hat{\w}^\t]$ is the noise-free estimation of $[\u^0, \w^0]$, $[\cdot,\cdot]$ denotes vector concatenation, and $\eta_k$ and $\lambda_k$ are denoising schedules. The final sampling step yields the solution $\w^0$ for the optimization problem in Eq. \ref{eq:joint_optimization}.

\section{Method}
\begin{algorithm}[ht]
    \small
    % \scriptsize
    \caption{Algorithm of \proj}
    \label{alg_total}
    \begin{algorithmic}[1]
    %  \SetAlgoLined
    \STATE \textbf{Require} Training set $D_{\textrm{train}}$, Calibration set $D_{\textrm{cal}}$, control targets\
    \STATE $\theta$ $\gets$ Pre-train with $\mathcal{L}_\textrm{diffusion}$ on $D_{\textrm{train}}$ \\
    \STATE $\theta$ $\gets$ Post-train with $\mathcal{L}_\textrm{post-train}$ on $D_{\textrm{train}}$ and $D_{\text{cal}}$ (Sec \ref{sec:posttrain}, Alg \ref{algpost}) \\
    \STATE $\theta, \w$ $\gets$ Fine-tune during guided inference with $\G$ and $\mathcal{L}_\textrm{fine-tune}$ on control targets and $D_{\textrm{cal}}$ (Sec \ref{sec:finetune}, Alg \ref{alg})
\hspace{0.8cm} \\
    \STATE \textbf{return} model parameters $\theta$ and control sequences $\w$
    \end{algorithmic}
\end{algorithm}

In this section, we introduce our proposed method \proj, with its overall framework outlined in Figure \ref{fig:overview} and Algorithm \ref{alg_total}. Firstly, in Section \ref{sec:conformal}, we propose the uncertainty quantile, which is the quantification of uncertainties. As shown in the middle of the figure, it is embedded throughout the algorithm, originating from the concern that the model's predictive uncertainty may cause a gap between the actual and predicted safety scores, potentially leading to unsafe events not anticipated by the model.

Secondly, in Section \ref{sec:posttrain} and \ref{sec:finetune}, we introduce the post-training and inference-time fine-tuning based on the uncertainty quantile, respectively. In the post-training phase following pre-training, we employ a reweighted loss function to guide the model's output distribution to favor regions with more optimal objectives and greater safety in general, after which the post-trained model can subsequently be used for specific control tasks. As for inference-time fine-tuning, we aim to make task-specific adjustments based on the post-trained model, enabling better control performance and safety guarantee for the specific control tasks with few iterations.

\subsection{Uncertainty Quantification of Diffusion Models}
\label{sec:conformal}

In offline safe control problems \cite{xu2022constraints, gu2022review}, the gap between pre-collected data and the target distribution exacerbates models' prediction errors, which can be critical in ensuring safety. To address this issue, we employ the conformal prediction technique to obtain uncertainty quantification for predicted safety, without requiring additional assumptions about the model and the data distribution.

The intuition behind the original conformal prediction \cite{vovk2005algorithmic} is to use the trained model on the calibration set to obtain a score set. The score set is then used to endow the model's prediction with a \emph{conformal (confidence) interval}. Under the assumption of exchangeability\footnote{Exchangeability means that exchanging examples in the calibration set and in inference does not alter their joint distribution.}, the actual prediction during inference will lie within the conformal interval with a guaranteed probability. The details of conformal prediction are provided in Appendix \ref{app:cp}. 

However, there is typically a distribution shift between the calibration set and the sequences generated by the diffusion model for control tasks during inference. Therefore, we account for this distribution shift, thereby introducing the \textit{shifted score set}. Based on it, we get the uncertainty quantification metric named \textit{uncertainty quantile} and finally the \textit{conformal interval}. The detailed description of this process is as follows.

% In the original version of conformal prediction \cite{vovk2005algorithmic}, the authors use the model on a subset of the training dataset, termed the \textit{calibration set}, to get a score set. Compared to this, we account for the distribution shift between the training dataset and the sequences generated by the diffusion model for control tasks, thereby introducing the \textit{shifted score set}. Based on it, we finally get the uncertainty quantification metric named \textit{uncertainty quantile}. The detailed description of this process is as follows.

\textbf{Calibration set and pre-training.} To achieve the goal mentioned above, we first set aside a portion of the original training data as the \textit{calibration set} \( D_{\text{cal}} \), which will be used later to estimate the model's prediction errors. The remaining data, which will be applied to operations that can alter the model’s outcomes, is referred to as the training data \( D_{\text{train}} \). After pre-training with \( D_{\text{train}} \) as described in Eq. \ref{eq:training_obj}, we get the diffusion model $p_\theta$ which models the joint distribution of $[\u,\w]$. 

\textbf{Shifted score set under distribution shift.} \textit{Score set}, obtained on the calibration set, is a set of model prediction uncertainties regarding the safety score $\score$, which is defined as 
\begin{equation}
    \S\coloneqq\{|\score(\pu(\w_i)) - \score(\u_i)|: (\u_i, \w_i)\in D_{\text{cal}}\}\cup\{\infty\}.
    \label{eq:scoreset}
\end{equation}
Here $\u_i$ and $\w_i$ are the system state trajectory and control signal sequence of the $i^\text{th}$ example, $\pu(\w)$ is the system state trajectory conditioned on control $\w_i$ as predicted by the model, where $\theta$ is the model parameter.

Furthermore, taking into account the distribution shift between the calibration set $D_\text{cal}$ and data generated based on control targets, we apply weighting to the score set $\S$ to get the \textit{shifted score set} as
\begin{equation}
\scalebox{1}{$
    \Tilde{\S}\coloneqq\{\omega_{\textrm{norm}}(\u_i,\w_i)\Delta\score_i: \Delta\score_i\in \S\}.
    \label{eq:weightedscore}
$}
\end{equation}
Intuitively, this weighting is because each sample in the calibration set has a different probability of appearing in the final data distribution generated by the model. According to conformal prediction under covariate shift \cite{Tibshirani2019ConformalPU}, here the calculation of the weights is as follows:
\vspace{-5pt}
\begin{equation}
\omega(\u_i,\w_i)\coloneqq\frac{\Tilde{p}(\u_i,\w_i)}{p(\u_i,\w_i)}
\vspace{-5pt}
\end{equation}
where $p(\u, \w)$ is the probability density function of the calibration set and training set, and $\Tilde{p}(\u, \w)$ is the probability density function of the model-generated data during inference. $\omega_\textrm{norm}$ is the normalization of $\omega$ defined as
\vspace{-5pt}
\begin{equation}
    \omega_\textrm{norm}(\u_i,\w_i)=\frac{\omega(\u_i,\w_i)}{\sum_{(\u_i, \w_i)\in D_{\text{cal}}}{\omega(\u_j,\w_j)}}.
\vspace{-5pt}
\end{equation}
The specific calculation of $\omega_\textrm{norm}$ regarding \proj will be detailed in Section \ref{sec:posttrain}.

\textbf{Uncertainty quantile.} Based on the shifted score set $\Tilde{\S}$ (Eq. \ref{eq:weightedscore}), the \textit{uncertainty quantile} is $\text{Quantile}((1-\alpha)(1+\frac{1}{|D_\text{cal}|});\Tilde{\S})$\footnote{It means obtaining the element with $(1-\alpha)(1+\frac{1}{|D_\text{cal}|})$'th quantile from the shifted score set $\Tilde{\S}$.}, where $1-\alpha$ is the coverage probability and $|D_\text{cal}|$ is the cardinality of $D_\text{cal}$. The quantile represents the threshold below which a specific proportion of $\Tilde{\S}$ falls. The adjustment term $1+1/|D_\text{cal}|$ accounts for the finite size of the calibration set, ensuring valid coverage even with limited data. Hereafter, it is abbreviated as $Q(1-\alpha;\Tilde{\S})$.

We can then obtain the \textit{conformal interval} as:
\begin{align}
\vspace{-5pt}
\notag
\textrm{CI}_\theta(1-\alpha,D_\textrm{cal})\coloneqq[\score(\pu(\w))-Q(1-\alpha;\Tilde{\S}), \\ 
\score(\pu(\w))+Q(1-\alpha;\Tilde{\S})]. 
\vspace{-5pt}
\end{align}
With at least $1-\alpha$ probability, the true $\score$ is covered by this interval, which can be demonstrated through the following theorem.
\begin{theorem}
     Assume that samples in the calibration set $D_\textrm{cal}\sim p$ are independent, and the test set $(\u,\w)\sim \Tilde{p}$ is also independent with the calibration set. Assume $p$ is absolutely continuous with respect to $\Tilde{p}$, then
     \begin{align}
         \mathbb{P}(\score(\u)\in\textrm{CI}_\theta(1-\alpha,D_\textrm{cal})) \geq 1-\alpha.
     \end{align}
     (See Appendix \ref{app:theory} for proof.)
\end{theorem}

\subsection{Post-training with Reweighted Loss}
\label{sec:posttrain}

The aim of post-training is to transform the model’s distribution into $p^*(\u,\w)\propto p(\u,\w) \cdot e^{-\mathcal{W}(\u,\w)}$, where $\W(\u, \w)$ contains both the control objective $\J$ and the safety constraints involving the model's uncertainty. Specifically, it is formulated as:
\begin{equation}
\scalebox{0.97}{$ \mathcal{W}(\u,\w) = \max[\score(\u)+Q(1-\alpha;\Tilde{\S})-\score_0, 0] + \gamma\J(\u,\w), $}
    \label{eq:w}
\end{equation}
where $\gamma$ is the weight of $\J$.

So we propose the reweighted post-training loss as 
\begin{align}
&\mathcal{L_\textrm{post-train}}\coloneqq\mathbb{E}_{\t\sim U(1,\T),(\u,\w)\sim p(\u,\w),\bepsilon\sim \mathcal{N}(\mathbf{0},\mathbf{I})} \notag \\ 
&[e^{-\mathcal{W}(\u,\w)}\|\mathbf{\bepsilon} - \bepsilon_\theta(\sqrt{\bar{\alpha}_\t}[\u,\w] + \sqrt{1-\bar{\alpha}_\t} \bepsilon,\t)\|_2^2],
\label{eq:posttrain_obj}
\end{align}
which steers the generated data distribution towards high $e^{-\mathcal{W}}$ regions. Compared to the pre-training loss in Eq. \ref{eq:training_obj}, the samples are reweighted, assigning higher weights to those with lower $\mathcal{W}$. This loss function enables the model to adjust the generated data distribution to align with our target distribution, as described in the following theorem:
\begin{theorem}
    Trained with $\mathcal{L_\textrm{post-train}}$ in Eq. \ref{eq:posttrain_obj}, the diffusion model's distribution is $p^*(\u,\w)\propto p(\u,\w) e^{-\mathcal{W}(\u,\w)}.$\\
    (See Appendix \ref{app:theory} for proof.)
\end{theorem}

Specifically, in the calculation of the uncertainty quantile $Q$, $\omega(\u_i,\w_i)=Cp_\theta(\u_i,\w_i)e^{-\mathcal{W}(\u_i,\w_i)}/p(\u_i,\w_i)$. In practical implementation, we let 
\begin{align}
    \omega(\u_i,\w_i)=Ce^{-\mathcal{W}(\u_i,\w_i)},
\end{align}
and the constant $C$ will be eliminated during normalization. Through the following theorem, we demonstrate that when the KL divergence between $p$ and $p_\theta$ is small, this approximation does not affect the results of the uncertainty quantile.
\begin{theorem}
    If $D_{KL}(p||p_\theta)\leq\epsilon$, then with $1-C\sqrt{\epsilon}$ probability, $\textrm{Quantile}(1-\alpha;\sum_{i=1}^N w_{\textrm{\textrm{norm}},i}\delta_{v_i})=\textrm{Quantile}(1-\alpha;\sum_{i=1}^N \tilde{w}_{\textrm{norm},i}\delta_{v_i})$, where $w_{\textrm{norm},i}$ and $\tilde{w}_{\textrm{norm},i}$ are the normalized $w_i$ and $\Tilde{w}_i$, $w_i=p_\theta(v_i) e^{-\mathcal{G}(v_i)}/p(v_i)$, and $\tilde{w}_i=e^{-\mathcal{G}(v_i)}$.\\
    (See Appendix \ref{app:theory} for proof.)
\end{theorem}

The algorithm of post-training is provided in Algorithm \ref{algpost} in Appendix \ref{app:algpost}.

\subsection{Inference-time Fine-tuning}
\label{sec:finetune}

During inference, knowing the specific control tasks, we guide the model to further optimize the generation of control sequences tailored to these tasks. We cyclically use the guidance to generate samples and then fine-tune the model parameters with these samples, thereby iteratively improving the model's safety and performance. In this process, similar to post-training, we incorporate the uncertainty quantile $Q$ into both the guidance and fine-tuning loss to prevent violations of safety constraints caused by the model’s uncertainty.

\textbf{Guidance $\G$.} Guidance is the first approach we adopt to steer the model’s output toward satisfying both the control objectives and safety constraints. It plays a role during the sampling process of the diffusion model. The specific denoising step of implementing guidance follows Eq. \ref{eq:1ddpm_inference}, and the form of guidance still follows $\mathcal{W}$ in Eq. \ref{eq:w}, \emph{i.e.} 
\begin{align}
    \G(\u,\w) = \W(\u,\w),
    \label{eq:guidance}
\end{align}
incorporating the uncertainty quantile $Q$ to account for safety risks arising from uncertainty.

\textbf{Fine-tuning.} The second approach for adjusting the output data distribution of the model is fine-tuning, which achieves the adjustment by optimizing the model parameters $\theta$. Specifically, retaining the computation graph for all denoising steps with respect to $\theta$ would result in an unmanageable memory overhead. Therefore, when we need to keep the computation graph during denoising for gradient calculation, we only retain the computation graph of the final denoising step. To optimize the safety score and  objective simultaneously, we form the fine-tune loss as:
\begin{equation}
\begin{aligned}
    &\mathcal{L}_{\text{fine-tune}} = \sum_{(\u_\theta,\w_\theta)\in D_{\text{sampled}}}\W(\u_\theta,\w_\theta),
    \label{eq:finetune}
\end{aligned}
\end{equation}
where $\u_\theta$ and $\w_\theta$ are from $D_\text{sampled}$ sampled according to the guidance described above.

The entire algorithm of inference-time finetuning is provided in Algorithm \ref{alg} in Appendix \ref{app:alg}.

\section{Experiment}
We design three safe PDE control scenarios, including the 1D Burgers’ equation, 2D incompressible fluid, and controlled nuclear fusion problem. These experiments aim to address the following three questions: (1) Does introducing the uncertainty quantile enable \proj to meet safety requirements? (2) Can \proj achieve superior control performance under safety constraints? (3) Do all components of our proposed algorithm contribute effectively to its performance?

For comparison, we choose imitation learning method Behavior Cloning (\textit{BC}) \citep{pomerleau1988alvinn}, and safe reinforcement learning and imitation learning methods involving \textit{BC} with safe data filtering (\textit{BC-Safe}), Constrained Decision Transformer (\textit{CDT}) \citep{liu2023constrained}
%and \textit{CDT} with safe data filtering (\textit{CDT-Safe})
and diffusion-based method \textit{TREBI} \citep{lin2023safe}. Note that \textit{CDT} shows the best performance in the offline Safe RL benchmark OSRL \citep{liu2023datasets}. In addition, we combine the physical system control method Supervised Learning \citep{hwang2022solving} and Model Predictive Control \citep{schwenzer2021review} with the Lagrangian approach \citep{chow2018risk} (\textit{SL-Lag, MPC-Lag}) to enforce safety constraints. We also apply the classical control method \textit{PID} \citep{1580152}. We provide the code \href{https://github.com/AI4Science-WestlakeU/safediffcon}{here}.
% https://anonymous.4open.science/r/Safe-Diffusion-Models-for-PDE-Control-213C/README.md

\begin{table}[ht]
\centering
\caption{\textbf{Results of 1D Burgers' equation.} {\color{gray} Gray}: there are unsafe trajectories. Black: all trajectories are safe. \textbf{Bold}: safe trajectories with \emph{lowest} $\J$.}
\vspace{-5pt}  % this might break margins, but worth trying.
\begin{tabular}{@{}l|c|ccc@{}}
\toprule
Methods    & $\J$ $\downarrow$ & $\R_{\text{sample}}$ $\downarrow$ & $\R_{\text{time}}$ $\downarrow$ & $\R_{\text{point}}$ $\downarrow$ \\ \midrule
BC & {\color{gray} 0.0001} & {\color{gray} 38\%} & {\color{gray} 13\%}   & {\color{gray} 1.2\%}  \\
BC-Safe                                               & {\color{gray} 0.0002}                    & {\color{gray} 14\%} & {\color{gray} 3\%}    & {\color{gray} 0.2\%} \\
PID                                                   & 0.0968       & 0\%                           & 0\%                             & 0.0\%                             \\
SL-Lag                  & 0.0115                                 & 0\%                           & 0\%                           & 0.0\%                             \\ 
MPC-Lag                  & 0.0092  & 0\%                           & 0\%                           & 0.0\%                             \\ 
CDT         & {\color{gray} 0.0012}     & {\color{gray} 16\%}   & {\color{gray} 3\%}      & {\color{gray} 0.2\%}                             \\
% CDT-Safe                          & 0.0021    & 0\%                           & 0\%                             & 0.0\%                             \\
TREBI   & 0.0074     & 0\%  & 0\%  & 0.0\%  \\
\midrule
\textbf{\proj} & {\textbf{0.0011}} & 0\%                           & 0\%                             & 0.0\%                             \\ \bottomrule
\end{tabular}
\label{table:1d}
\end{table}

\begin{figure}[t]
\vspace{-2pt}  % this might break margins, but worth trying.
\begin{center}
    \includegraphics[width=\columnwidth]{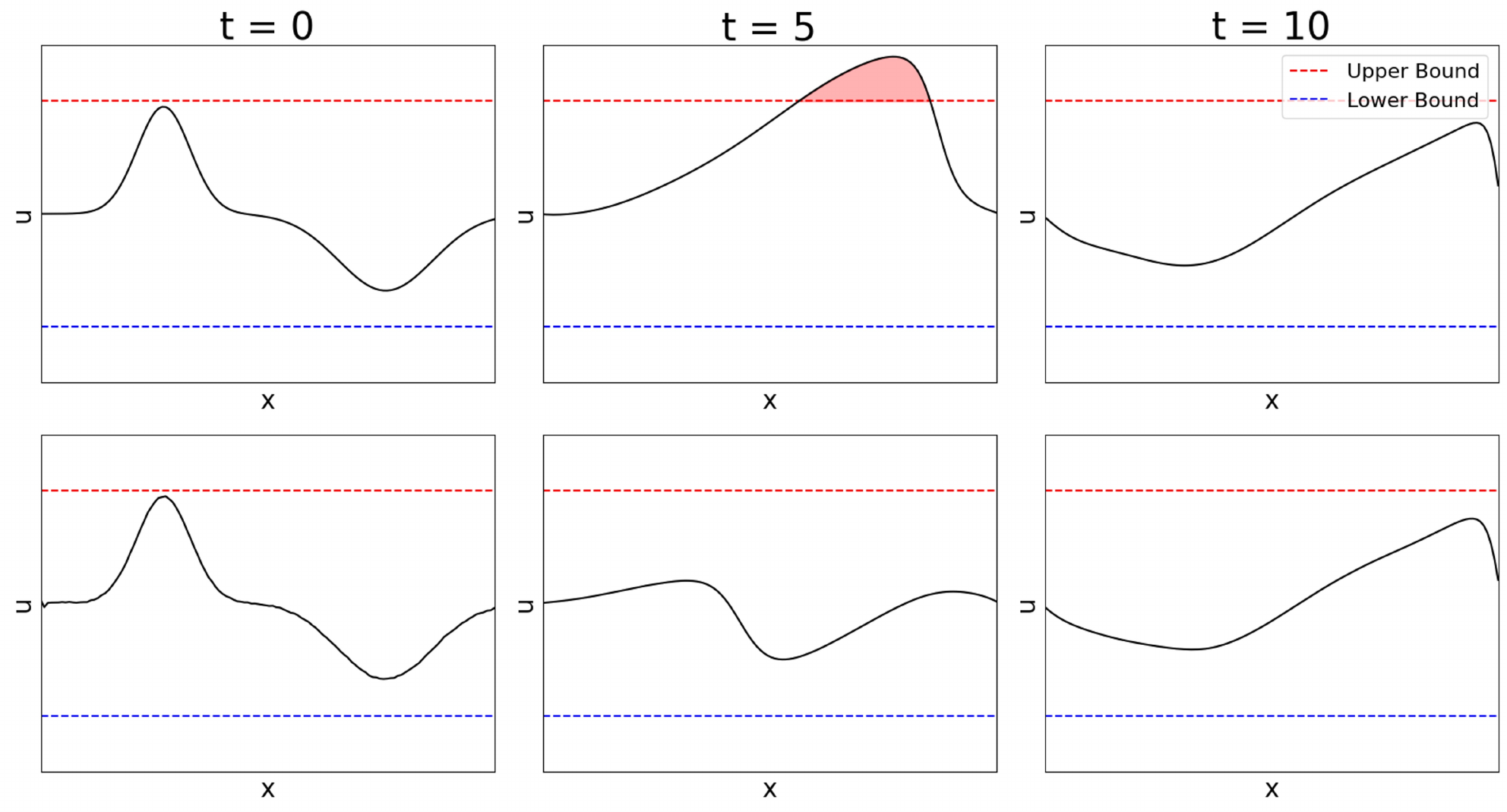}
\end{center}
\vspace{-8pt}
\caption{\textbf{Visualizations of the 1D Burgers' equation.} The top row shows the original trajectory corresponding to the control target, and the bottom row is the trajectory controlled by \proj.}
\vspace{-6pt}  % this might break margins, but worth trying.
\label{fig:1d}
\end{figure}

\subsection{1D Burgers' Equation}
\label{sec:1d}
\textbf{Experiment settings.}
1D Burgers' equation is a fundamental equation that governs various physical systems including fluid dynamics and gas dynamics. Here we follow previous works \citep{hwang2022solving,mowlavi2023optimal} and consider the Dirichlet boundary condition along with an external force \(\w (t,x) \). 

Given a target state $\u_d(x)$, the control objective $\J$ is to minimize the control error between the final state $\u_T$ and the target state $\u_d$.
\begin{equation}
\label{eq:burgers_obj_J_actual}
\J\coloneqq\int_{\Omega}|\u(T,x) - \u_d(x)|^2\mathrm{d}x.
\end{equation}
The safety score is defined as:
\begin{equation}
    s(\u) \coloneqq \sup_{(t,x) \in [0,T] \times \Omega}\{\u(t,x)^2\},
\end{equation}\label{eq:burgers_safety_score}
with the safety constraint $s_0$. If $s(\u)>s_0$, the state trajectory $\u$ is unsafe, otherwise, it is safe. We compute three unsafe rates to assess the safety levels of different methods' control results. $\R_{\text{sample}}$ denotes the proportion of unsafe trajectories among total trajectories\footnote{If any point in the full trajectory is unsafe, this trajectory is unsafe. So $\R_{\text{sample}}$ is the most stringent metric.}; $\R_{\text{time}}$ denotes the proportion of unsafe timesteps among all timesteps; $\R_{\text{point}}$ denotes the proportion of unsafe spatial lattice points in all spatial lattice points across all time steps. More details can be found in Appendix \ref{app:1dexp}. 

% To better evaluate whether the set of state trajectories controlled by the model is safe, we define the normalized safety score:
% \begin{equation}
%     \score_{\text{norm}} \coloneqq \frac{1}{|\mathcal{N}_1|} \sum_{i \in \mathcal{N}_1} \frac{s(\u_i)}{s_0} +\frac{1}{|\mathcal{N}_2|} \sum_{i \in \mathcal{N}_2} \frac{s(\u_i)}{s_0},
% \end{equation}
% where $\mathcal{N}_1 = \{ i \mid \u_i \leq s_0\}$, $\mathcal{N}_2 = \{ i \mid \u_i > s_0 \}$. Note that $s(\u)$ and $s_0$ are always non-negative. If the state trajectories are all safe, the score is smaller than 1; If \emph{any} state trajectory is unsafe, the cost is greater than 1. Therefore, when $\score_{\text{norm}}$ is less than 1, the different algorithms only need to compare the control objective $\J$.

\textbf{Results.} In Table \ref{table:1d}, We report the results of the control objective $\J$ and safety metrics of different methods. \proj can meet the safety constraint and achieve the best control objective at the same time. As shown in Figure \ref{fig:1d}, given the initial condition and the final state (control target), \proj can control a state trajectory that satisfies the safety constraint and control target. Other methods either suffer from constraint violations or suboptimal objectives. \textit{BC} and \textit{BC-Safe} trained from expert trajectories fail to meet the safety constraints, showing that simple behavior cloning is not feasible under the safety constraints. \textit{SL-Lag} and \textit{MPC-Lag} attempt to use the Lagrangian method to balance the control objective and safety, but this coupled training program makes it difficult to find the right balance, with poor control performance. \textit{CDT} uses the complex Transformer architecture, which can achieve low control error, but it can not
% needs to filter unsafe data (\textit{CDT-Safe}) to 
meet safety constraints. The diffusion-based method \textit{TREBI} sacrifices too much control performance to satisfy the safety constraints, because its error bound is soft, and the safety constraints can be easily violated.

\subsection{2D Incompressible Fluid}
\begin{figure*}[t]
% \vspace{-5pt}  % this might break margins, but worth trying.
\begin{center}
    \includegraphics[scale=0.35]{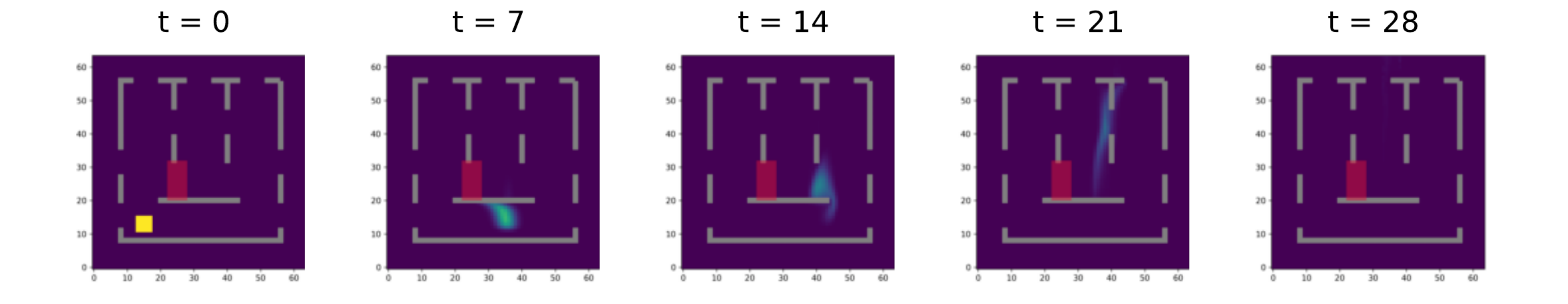}
\end{center}
\vspace{-13pt}
\caption{\textbf{Visualization of the 2D incompressible fluid control by our \proj.} By controlling the fluid on the outside margin, the yellow smoke is successfully maneuvered to the center top exit while avoiding the red unsafe region.}
\vspace{-13pt}
\label{fig:2d}
\end{figure*}

% \begin{table}[ht]
% \centering
% % \vspace{-10pt}  % this might break margins, but worth trying.
% \caption{\textbf{2D incompressible fluid control results.} {\color{gray} Gray}: $\score_{\text{norm}}$ is greater than 1. Black: $\score_{\text{norm}}$ is smaller than 1. \textbf{Bold}: methods marked in black with lowest $\J$.}
% \begin{tabular}{@{}l|rr|cr@{}}
%     \toprule
%     Methods    & $\J$ $\downarrow$  & $\score_\text{norm}$ $\downarrow$ & $\max[\score-\score_0, 0]$ $\downarrow$ & $\R$ $\downarrow$ \\ \midrule
%     BC  &  \color{gray}{-0.7125}  &  \color{gray}7.3402  & \color{gray}0.7160 &  \color{gray}88\%  \\
%     BC-Safe &  {-0.2520}  &  {0.3463}  &   {0.0330} &  {8\%}   \\
%     CDT     &  \color{gray}-0.7133  &  \color{gray}3.0778  &   \color{gray}0.2726 &  \color{gray}34\%  \\
%     CDT-Safe  & {-0.6360} &  {0.5073}  &   {0.0292} &  {18\%}  \\ \midrule
%     ${\textbf{\proj (Ours)}}$   & {\textbf{-0.7035}}   & {\textbf{0.6092}}    & {\textbf{0.0380}}   & {\textbf{14\%}}   \\ \bottomrule
% \end{tabular}
% \label{tab:2d}
% \end{table}

\textbf{Experiment settings.} We then consider the control problems of the 2D incompressible fluid, which follows the Navier-Stokes equation. Following previous works \citep{holl2020learning, wei2024generative, hu2024wavelet}, the control task we consider is to maximize the amount of smoke that passes through the target bucket in the fluid flow with obstacles and openings, while constraining the amount of smoke passing through the dangerous region under the safety bound. Specifically, referring to Figure \ref{fig:2d}, the control objective \( \J \) is defined as the negative ratio of smoke passing through the target bucket located at the center top, while the safety score \( \score \) corresponds to the ratio of smoke entering the hazardous red region. It is important to note that there is a trade-off between controlling the flow through the hazardous region and achieving a more optimal control objective, which imposes higher demands on the algorithm. We set the safety score bound as \( \score_0 = 0.1 \).

Moreover, this control task is particularly challenging due to its specific setup: not only does it require indirect control, which means that control can only be applied to the peripheral region, but the spatial control parameters also reach as many as 1,792. As for safety, among all the training data, 53.1\% of the samples are unsafe, meaning their safety score \(\score\) exceeds the bound \(\score_0 = 0.1\). The average safety score of the dataset is 0.3215. Other details can be found in Appendix \ref{app:2dexp}.

\begin{table}[ht]
\centering
\vspace{-2pt}  % this might break margins, but worth trying.
\caption{\textbf{2D incompressible fluid control results.} {\color{gray} Gray}: there are unsafe trajectories. Black: all trajectories are safe. \textbf{Bold}: safe trajectories with \emph{lowest} $\J$.}
\vspace{-5pt}  % this might break margins, but worth trying.
\begin{tabular}{@{}l|c|cc@{}}
    \toprule
    Methods    & $\J$ $\downarrow$ & $SVM$ $\downarrow$ & $\R$ $\downarrow$ \\ \midrule
    BC  &  \color{gray}{-0.7104}  & \color{gray}0.7156 &  \color{gray}88\%  \\
    BC-Safe &  \color{gray}{-0.2520}  &   \color{gray}{0.0330} &  \color{gray}{8\%}   \\
    CDT     &  \color{gray}-0.7025  &   \color{gray}0.2519 &  \color{gray}30\%  \\
    % CDT-Safe  & \color{gray}{-0.6391}  &   \color{gray}{0.0158} &  \color{gray}{16\%}  \\ 
    TREBI   &  \color{gray}-0.7019  & \color{gray}0.0808  &  \color{gray}18\%  \\\midrule
    ${\textbf{\proj}}$   & {\textbf{-0.3548}}  & {{0.0000}}   & {{0\%}}   \\ \bottomrule
\end{tabular}
\vspace{-2pt}  % this might break margins, but worth trying.
\label{tab:2d}
\end{table}

\textbf{Results.} We report results of \proj and baselines in Table \ref{tab:2d}. Here \textit{SL-Lag} and \textit{MPC-Lag} fail to achieve reasonable control results because of the complex dynamics, and we define $\R$ as the rate of unsafe samples. Additionally, we introduce another metric called Safety Violation Magnitude $SVM=\max[\score - \score_0, 0]$. When \(\score\) does not exceed the bound \(\score_0\), this metric is 0. If \(\score\) exceeds \(\score_0\), the metric reflects the amount by which it is surpassed. From the results, it is clear that our method is the only one to successfully satisfy the safety constraint, even surpassing methods that only consider safety like BC-Safe, which demonstrates the effectiveness of post-training and fine-tuning with the uncertainty quantile.

\subsection{Tokamak Fusion Reactor}
\textbf{Experiment settings.}
A tokamak is a device designed to achieve controlled nuclear fusion by confining high-temperature plasma within a toroidal magnetic field, enabling conditions suitable for fusion reactions \cite{federici2001plasma}. The realization of controlled fusion could provide humanity with a clean, safe, and nearly limitless energy source, free from greenhouse gases, nuclear accident risks, and fuel supply imbalances \cite{schuster2006role}. Safety in a tokamak is crucial to prevent plasma instabilities and disruptions \cite{kates2019predicting}, which can damage the device worth billions of dollars.

Controlling the plasma state in a tokamak is critical for maintaining stability and optimizing fusion performance, which often requires precise management of parameters like beta ($\beta_p$), safety factor ($q_{95}$), and internal inductance ($l_i$). Based on the previous works \citep{seo2021feedforward, seo2022development}, we set the control objective to make $\beta_p$ and $l_i$ reach specific target values $\beta_p^*$ and $l_i^*$ at each time step, while the safety constraint requires $q_{95}$ to remain above the defined safety bound at all times. In this case, the objective $\J$ and safety score $\score$ are formulated as follows:
\begin{align}
    &\J\coloneqq\int_{\Omega\times[0,T]}(|\beta_p(t,x) - \beta_p^*(x)|^2 \\
    &\quad\quad\quad\quad\quad+|l_i(t,x) - l_i^*(x)|^2)\mathrm{d}x\mathrm{d}t, \\
    &\score\coloneqq-\inf_{(t,x) \in [0,T] \times \Omega}\{q_{95}(t,x)\}.
\end{align}
And the safety score bound is $\score_0=-4.98$. We use the same initial state and tokamak simulation environment as in the previous work \citep{seo2022development}, setting time-varying random targets within the reasonable ranges proposed in that study. The control sequences for training are collected using the reinforcement learning model trained in the previous work \citep{seo2022development}, where 71.18\% of the samples in the training dataset are unsafe.

\begin{table}[ht]
\centering
\vspace{-4pt}  % this might break margins, but worth trying.
\caption{\textbf{Results of the tokamak fusion reactor.} {\color{gray} Gray}: there are unsafe trajectories. Black: all trajectories are safe. \textbf{Bold}: safe trajectories with \emph{lowest} $\J$.}
\vspace{-3pt}  % this might break margins, but worth trying.
\begin{tabular}{@{}l|c|cc@{}}
\toprule
Methods    & $\J$ $\downarrow$   & $\R_{\text{sample}}$ $\downarrow$ & $\R_{\text{time}}$ $\downarrow$ \\ \midrule
BC                  & \color{gray}0.0610   &  \color{gray}42\%  &  \color{gray}1.34\%  \\
BC-Safe             & \color{gray}0.0811   &   \color{gray}4\%  &  \color{gray}0.03\%  \\
SL-Lag              & 0.8812   &   0\%  &  0.00\%  \\ 
MPC-Lag             & 0.8659   &   0\%  &  0.00\%  \\ 
CDT                 & \color{gray}0.0071   &   \color{gray}8\%  &  \color{gray}0.54\%  \\
% CDT-Safe               &    &    &  \\
TREBI               & 0.0261   &   0\%  &  0.00\%  \\
\midrule
\textbf{\proj}  & {\color[HTML]{000000} \textbf{0.0094}} & {0\%} & {0.00\%} \\ \bottomrule
\end{tabular}
\label{table:tokamak}
\end{table}
\vspace{-2pt}  % this might break margins, but worth trying.

\textbf{Results.}
We provide the results of control and safety metrics in Table \ref{table:tokamak}, where the definition of $\R_{\text{sample}}$ and $\R_{\text{time}}$ are the same as Section \ref{sec:1d}. Due to the differences in the input and output dimensions of the controller and the lack of clarity regarding the device's specific details, PID cannot be directly applied here. Results show that \proj successfully achieves full compliance with safety constraints. Moreover, among all algorithms that satisfy safety constraints, our algorithm achieves the lowest $\J$, indicating that it also achieves the best control performance. These results verify our statement that our algorithm is not only safe but also effective.

\subsection{Ablation Study}
To demonstrate the necessity and effectiveness of each component of our algorithm, we conduct three ablation experiments on the 2D incompressible fluid control problem by removing post-training, inference-time fine-tuning, and the uncertainty quantile $Q$, respectively.

It can be observed that post-training and fine-tuning, as two methods that influence the model’s output distribution, are both essential for meeting safety constraints. Removing either one results in the algorithm failing to achieve safety, highlighting the indispensable roles of these two components. Furthermore, when the uncertainty quantile is removed, which means $Q$ is set to 0, the algorithm also fails to meet safety requirements. This demonstrates that without considering the model’s uncertainty, its safety predictions are inaccurate, leading to the occurrence of unsafe events.

% \begin{table}[ht]
% \centering
% \caption{\textbf{Results of the ablation study.}}
% \begin{tabular}{@{}l|ccc|ccc@{}}
%     \toprule
%           & \multicolumn{3}{c|}{1D}      & \multicolumn{3}{c}{2D}       \\
%           & \proj  & w/o $\G$ & w/o $\mathcal{L}_{\text{fine-tune}}$ & \proj  & w/o $\G$ & w/o $\mathcal{L}_{\text{fine-tune}}$ \\ \midrule
%     $\J$                 &    &    &    &    &    &     \\
%     $\score_\text{norm}$ &    &    &    &    &    &   \\ \bottomrule
% \end{tabular}
% \label{tab:ablation}
% \end{table}

\begin{table}[ht]
\vspace{-4pt}  % this might break margins, but worth trying.
\centering
\caption{\textbf{Results of the ablation study.} We compare \proj with \proj w/o post-training, inference-time fine-tuning and the uncertainty quantile $Q$.}
\vspace{-3pt}  % this might break margins, but worth trying.
\begin{tabular}{@{}l|c|cc@{}}
    \toprule
    Methods    & $\J$ $\downarrow$ & $SVM$ $\downarrow$ & $\R$ $\downarrow$ \\ \midrule
    \proj             &  -0.3548  &  0.0000  &  0\%  \\ 
    w/o post-training &  -0.4911  &  0.0849  &  12\%  \\
    w/o fine-tuning   &  -0.4692  &  0.0247  &  8\%  \\ 
    w/o $Q$           &  -0.6364  &  0.0623  &  12\%  \\ \bottomrule
\end{tabular}
\vskip -6pt
\label{tab:ablation}
\end{table}

% We highlight that one key distinction between our framework and previous safe RL methods lies in the introduction of fine-tuning within Iterative Safety Improvement, which updates the model parameters based on specific control tasks and safety constraints. To further validate the effectiveness of our proposed Iterative Safety Improvement, we conduct experiments using a version of \proj without the fine-tuning component. As shown in the table, without fine-tuning, \proj exhibits a significant decline in safety, both in 1D and 2D settings, with the 1D case becoming notably unsafe. This emphasizes the importance and effectiveness of the Iterative Safety Improvement framework in addressing safety control problems.
\section{Limitation and Future Work}
Firstly, our method is not constrained by specific scenarios, meaning it can be applied to more real-world tasks, which is our future work. Secondly, we consider extending this method to other generative methods that require constraints, not just the diffusion model. Finally, in the future, we will explore the possibility of developing a stricter coverage of the safety score that sacrifices less accuracy while still ensuring safety.

\section{Conclusion}
In this paper, we have introduced \emph{\underline{Safe} \underline{Diff}usion Models for PDE \underline{Con}trol} (\proj), a novel approach that integrates safety constraints into deep learning-based control of PDE systems. By quantifying safety uncertainty using conformal prediction and introducing a reweighted loss during post-training, our method effectively handles distribution shifts and ensures safety while maintaining competitive control performance. Furthermore, inference-time fine-tuning allows dynamic adjustment of the model to meet specific control targets. The results of experiments on diverse control tasks, including 1D Burgers’ equation, 2D incompressible fluid, and controlled nuclear fusion, demonstrate that \proj reliably satisfies safety constraints and outperforms existing methods. This work paves the way for safer deployment of machine learning techniques in high-stakes real-world applications.

% % Acknowledgements should only appear in the accepted version.
% \section*{Acknowledgements}

% \textbf{Do not} include acknowledgements in the initial version of
% the paper submitted for blind review.

% If a paper is accepted, the final camera-ready version can (and
% usually should) include acknowledgements.  Such acknowledgements
% should be placed at the end of the section, in an unnumbered section
% that does not count towards the paper page limit. Typically, this will 
% include thanks to reviewers who gave useful comments, to colleagues 
% who contributed to the ideas, and to funding agencies and corporate 
% sponsors that provided financial support.

\section*{Impact Statement}
This paper presents work whose goal is to advance the field of Machine Learning. Furthermore, our work makes Machine Learning applications safer in real-world scenarios. There are many potential societal consequences of our work, none of which we feel must be specifically highlighted here.


\bibliography{main}
\bibliographystyle{icml2025}

%%%%%%%%%%%%%%%%%%%%%%%%%%%%%%%%%%%%%%%%%%%%%%%%%%%%%%%%%%%%%%%%%%%%%%%%%%%%%%%
%%%%%%%%%%%%%%%%%%%%%%%%%%%%%%%%%%%%%%%%%%%%%%%%%%%%%%%%%%%%%%%%%%%%%%%%%%%%%%%
% APPENDIX
%%%%%%%%%%%%%%%%%%%%%%%%%%%%%%%%%%%%%%%%%%%%%%%%%%%%%%%%%%%%%%%%%%%%%%%%%%%%%%%
%%%%%%%%%%%%%%%%%%%%%%%%%%%%%%%%%%%%%%%%%%%%%%%%%%%%%%%%%%%%%%%%%%%%%%%%%%%%%%%
\newpage
\appendix
\onecolumn

\section{Theory}
\label{app:theory}

In this section, we provide the proof of the theorems mentioned in the main text.

\begin{theorem}
     Assume samples in the calibration set $D_\textrm{cal}\sim p$ are independent, and the test set $(\u,\w)\sim \Tilde{p}$ is also independent with the calibration set. Assume $p$ is absolutely continuous with respect to $\Tilde{p}$, then
     \begin{align}
         \mathbb{P}(\score(\u)\in\textrm{CI}_\theta(1-\alpha,D_\textrm{cal})) \geq 1-\alpha.
     \end{align}
\end{theorem}

\paragraph{Proof.} According to Lemma 2 in the previous work \citep{Tibshirani2019ConformalPU}, the calibration set and test set are weighted exchangeable (defined in Definition 1 in the previous work \citep{Tibshirani2019ConformalPU}), with the weight function $\omega$.

According to Lemma 3 in the previous work \citep{Tibshirani2019ConformalPU}, let the $(\u,\w)$ pairs be the random variables, we can obtain that 
\begin{align}
 \mathbb{P}(\score(\u)\leq\score(\pu(\w))+Q(1-\alpha;\Tilde{\S})) \geq 1-\alpha.
\end{align}
Similarly, it can be derived that
\begin{align}
 \mathbb{P}(\score(\pu(\w))-Q(1-\alpha;\Tilde{\S})\leq\score(\u)) \geq 1-\alpha.
\end{align}
Thus      
\begin{align}
    \mathbb{P}(\score(\u)\in\textrm{CI}_\theta(1-\alpha,D_\textrm{cal})) \geq 1-\alpha.
\end{align} \qed

\begin{theorem}
    Trained with $\mathcal{L_\textrm{post-train}}$
    \begin{align}
    \mathcal{L_\textrm{post-train}}=\mathbb{E}_{\t\sim U(1,\T),(\u,\w)\sim p(\u,\w),\bepsilon\sim \mathcal{N}(\mathbf{0},\mathbf{I})}
    [e^{-\mathcal{W}(\u,\w)}\|\mathbf{\bepsilon} - \bepsilon_\theta(\sqrt{\bar{\alpha}_\t}[\u,\w] + \sqrt{1-\bar{\alpha}_\t} \bepsilon,\t)\|_2^2],
    \end{align}
    the diffusion model's distribution is $p^*(\u,\w)\propto p(\u,\w) e^{-\mathcal{W}(\u,\w)}.$
\end{theorem}

\paragraph{Proof.} Trained with the diffusion loss
\begin{align}
\mathcal{L_\textrm{diffusion}}=\mathbb{E}_{\t\sim U(1,\T),(\u,\w)\sim p(\u,\w),\bepsilon\sim \mathcal{N}(\mathbf{0},\mathbf{I})}
[\|\mathbf{\bepsilon} - \bepsilon_\theta(\sqrt{\bar{\alpha}_\t}[\u,\w] + \sqrt{1-\bar{\alpha}_\t} \bepsilon,\t)\|_2^2],
\end{align}
the diffusion model's distribution is $p$, thus to learn the distribution $p^*$, the loss should be 
\begin{align}
\mathcal{L}&=\mathbb{E}_{\t\sim U(1,\T),(\u,\w)\sim p^*(\u,\w),\bepsilon\sim \mathcal{N}(\mathbf{0},\mathbf{I})}
[\|\mathbf{\bepsilon} - \bepsilon_\theta(\sqrt{\bar{\alpha}_\t}[\u,\w] + \sqrt{1-\bar{\alpha}_\t} \bepsilon,\t)\|_2^2] \\
&=\mathbb{E}_{\t\sim U(1,\T),\bepsilon\sim \mathcal{N}(\mathbf{0},\mathbf{I})}
[\int\|\mathbf{\bepsilon} - \bepsilon_\theta(\sqrt{\bar{\alpha}_\t}[\u,\w] + \sqrt{1-\bar{\alpha}_\t} \bepsilon,\t)\|_2^2 \cdot p^*(\u,\w) \textrm{d}\u\textrm{d}\w] \\
&\propto\mathbb{E}_{\t\sim U(1,\T),\bepsilon\sim \mathcal{N}(\mathbf{0},\mathbf{I})}
[\int\|\mathbf{\bepsilon} - \bepsilon_\theta(\sqrt{\bar{\alpha}_\t}[\u,\w] + \sqrt{1-\bar{\alpha}_\t} \bepsilon,\t)\|_2^2 \cdot p(\u,\w)e^{-\W(\u,\w)} \textrm{d}\u\textrm{d}\w] \\
&=\mathbb{E}_{\t\sim U(1,\T),(\u,\w)\sim p(\u,\w),\bepsilon\sim \mathcal{N}(\mathbf{0},\mathbf{I})}
[e^{-\W(\u,\w)}\|\mathbf{\bepsilon} - \bepsilon_\theta(\sqrt{\bar{\alpha}_\t}[\u,\w] + \sqrt{1-\bar{\alpha}_\t} \bepsilon,\t)\|_2^2].
\end{align} \qed

\begin{theorem}
    If the KL divergence of the real distribution $p$ and the distribution $p_\theta$ learned by the model is less than $\epsilon$, then with $1-C\sqrt{\epsilon}$ probability, the quantile $\textrm{Quantile}(1-\alpha;\sum_{i=1}^N w_{\textrm{\textrm{norm}},i}\delta_{v_i})=\textrm{Quantile}(1-\alpha;\sum_{i=1}^N \tilde{w}_{\textrm{norm},i}\delta_{v_i})$, where $w_{\textrm{norm},i}$ and $\tilde{w}_{\textrm{norm},i}$ are the normalized $w_i$ and $\Tilde{w}_i$, $w_i=\frac{p_\theta(v_i) e^{-\mathcal{W}(v_i)}}{p(v_i)}$ and $\tilde{w}_i=e^{-\mathcal{W}(v_i)}$.
\end{theorem}

\paragraph{Proof.} Since $D_{KL}(p||p_\theta)=\int p(x)\log\frac{p(x)}{p_\theta(x)}dx\leq\epsilon$, we can get
\begin{align}
\mathbb{P}(\mathcal{A}=\{|p(x)-p_\theta(x)|>\delta\})
&\leq\frac{\int_\mathcal{A}|p(x)-p_\theta(x)|dx}{\delta} \\
&\leq\frac{\int_\Omega|p(x)-p_\theta(x)|dx}{\delta}\\
&=\frac{2TV(p,p_\theta)}{\delta} \\
&\leq\frac{2}{\delta}\sqrt{\frac{1}{2}D_{KL}(p||p_\theta)} \label{eq:pinsker} \\
&\leq\frac{\sqrt{2\epsilon}}{\delta},
\end{align}
where $TV(p,p_\theta)$ is the total variation distance between $p$ and $p_\theta$. Eq. \ref{eq:pinsker} is obtained using Pinsker’s inequality. This inequality means that with $1-\frac{\sqrt{2\epsilon}}{\delta}$ probability, $|\Delta p(v)|:=|p_\theta(v)-p(v)|\leq\delta$.

Let $\delta$ be a small quantity and $p_-=\min_{i=1,\cdots,N}(p(v_i))$, since
\begin{align}
    \sum_j(1+\frac{|\Delta p(v_j)|}{p_-})\geq\sum_j(1+\frac{\Delta p(v_j)}{p_j})\geq\sum_j(1-\frac{|\Delta p(v_j)|}{p_-}),
\end{align}
and 
\begin{align}
    w_{\textrm{norm},i}=\frac{w_i}{\sum_{j=1}^N w_j}=\frac{(1+\frac{\Delta p(v_i)}{p(v_i)})\Tilde{w}_i}{\sum_{j=1}^N(1+\frac{\Delta p(v_j)}{p(v_j)})\Tilde{w}_j},
\end{align}
thus
\begin{align}
    \frac{1+\frac{\Delta p(v_i)}{p(v_i)}}{1+\frac{|\Delta p(v_i)|}{p_-}}\Tilde{w}_{\textrm{norm},i}\leq w_{\textrm{norm},i}\leq\frac{1+\frac{\Delta p(v_i)}{p(v_i)}}{1-\frac{|\Delta p(v_i)|}{p_-}}\Tilde{w}_{\textrm{norm},i}.
\end{align}
By rearranging terms, we obtain
\begin{align}
    |w_{\textrm{norm},i}-\Tilde{w}_{\textrm{norm},i}|
    &\leq\max\{|\frac{\Delta p(v_i)/p(v_i)-|\Delta p(v_i)|/p_-}{1+|\Delta p(v_i)|/p_-}|, |\frac{\Delta p(v_i)/p(v_i)+|\Delta p(v_i)|/p_-}{1-|\Delta p(v_i)|/p_-}| \} \\
    &\leq|\Delta p(v_i)|\frac{|1/p(v_i)|+|1/p_-|}{1-\delta/p_-} \\
    &\leq C|\Delta p(v_i)| \\
    &\leq C|\delta|. \label{eq:w_diff}
\end{align}
Because we can merge the $v_i$ values that are the same, we can assume $v_1<v_2<\cdots<v_N$. Suppose $\textrm{Quantile}(\alpha;\sum_{i=1}^N w_{\textrm{norm},i}\delta_{v_i})=v_k$, which means
\begin{align}
    \sum_{i=i}^k w_{\textrm{norm},i}\geq1-\alpha, \sum_{i=i}^{k-1} w_{\textrm{norm},i}<1-\alpha.
\end{align}
With Eq. \ref{eq:w_diff}, we can get that
\begin{align}
    \sum_{i=i}^k \Tilde{w}_{\textrm{norm},i}\geq1-\alpha-Ck\delta, \sum_{i=i}^{k-1} \Tilde{w}_{\textrm{norm},i}<1-\alpha+Ck\delta.
\end{align}
We can let $\delta$ be small enough to make $k$ satisfy
\begin{align}
    \sum_{i=i}^k \Tilde{w}_{\textrm{norm},i}\geq1-\alpha, \sum_{i=i}^{k-1} \Tilde{w}_{\textrm{norm},i}<1-\alpha,
\end{align}
which means $\textrm{Quantile}(\alpha;\sum_{i=1}^N w_{\textrm{norm},i}\delta_{v_i})=\textrm{Quantile}(\alpha;\sum_{i=1}^N \tilde{w}_{\textrm{norm},i}\delta_{v_i})$. \qed

\section{Additional Experiment Results}
\subsection{Inference Time}
We have accelerated the efficiency of inference. On one hand, by introducing post-training, the data distribution is already close to the desired distribution before inference. On the other hand, we have significantly accelerated the diffusion model's sampling time using DDIM \cite{song2021denoising}. The Table \ref{tab:time} below presents a comparison of the inference time between our method and the baseline on the Burgers' equation on A800 with 8 CPUs. It shows that our method achieves a moderate speed and is much faster compared to TREBI, which is also based on the diffusion model.

\begin{table}[ht]
\centering
\caption{\textbf{Inference time on the Burgers' equation on A800 with 8 CPUs.}}
\begin{tabular}{@{}l|ccccccc@{}}
\toprule
Methods &	BC &	PID &	SL-Lag &	MPC-Lag &	CDT &	TREBI &	\proj \\ \midrule
Inference Time (min) &	0.1351 &	0.1091 &	0.8842 &	26.2905 &	0.0890 &	13.5525 &	2.3575 
\\ \bottomrule
\end{tabular}
\label{tab:time}
\end{table}

\subsection{Ablation Study}
In this subsection, we provide results of ablation studies on the 1D Burgers' equation and the tokamak fusion reactor. We also compare \proj and \proj without post-training, inference-time fine-tuning, and the uncertainty quantile $Q$. The results in Table \ref{tab:ablation_burgers} and Table \ref{tab:ablation_tokamak} still show that the absence of any module affects the model, causing it to fail to meet the safety constraints. Therefore, each module is effective.

\begin{table}[ht]
\centering
\caption{\textbf{Results of the ablation study on the Burgers' equation.} We compare \proj with \proj w/o post-training, inference-time fine-tuning and the uncertainty quantile $Q$.}
\begin{tabular}{@{}l|c|ccc@{}}
\toprule
Methods    & $\J$ $\downarrow$ & $\R_{\text{sample}}$ $\downarrow$ & $\R_{\text{time}}$ $\downarrow$ & $\R_{\text{point}}$ $\downarrow$ \\ \midrule
    \proj             &   0.0011  &	 0\% &	0\%	&  0\%  \\ 
    w/o post-training &  0.0014 &	4\% &	50\% &	0.01\%  \\
    w/o fine-tuning   &  0.0007 &	40\% &	20\% &	3\%  \\ 
    w/o $Q$           &  0.0006 &	30\% &	10\% &	1\%  \\ \bottomrule
\end{tabular}
\label{tab:ablation_burgers}
\end{table}

\begin{table}[ht]
\centering
\caption{\textbf{Results of the ablation study on the tokamak fusion reactor.} We compare \proj with \proj w/o post-training, inference-time fine-tuning and the uncertainty quantile $Q$.}
\begin{tabular}{@{}l|c|cc@{}}
\toprule
Methods    & $\J$ $\downarrow$   & $\R_{\text{sample}}$ $\downarrow$ & $\R_{\text{time}}$ $\downarrow$ \\ \midrule
    \proj             &   0.0094 &	0\% &	0\%  \\ 
    w/o post-training &  0.0153 &	10\% &	0.52\%  \\
    w/o fine-tuning   &  0.0210 &	50\% &	7.74\%  \\ 
    w/o $Q$           &  0.0269 &	28\% &	0.65\%
  \\ \bottomrule
\end{tabular}
\label{tab:ablation_tokamak}
\end{table}

\subsection{Analysis on Parameters}
We conduct experimental analysis on coverage probability $\alpha$, the split ratio of training/calibration data, and the weight of objective $\gamma$. The Table \ref{tab:alpha}, \ref{tab:split_ratio} and \ref{tab:gamma} show that they do not impact much on performance, indicating that the model is robust.

\begin{table}[ht]
\centering
\caption{\textbf{Impact of the coverage probability $\alpha$ on the tokamak fusion reactor.}}
\begin{tabular}{@{}l|c|cc@{}}
\toprule
Split Ratio    & $\J$ $\downarrow$   & $\R_{\text{sample}}$ $\downarrow$ & $\R_{\text{time}}$ $\downarrow$ \\ \midrule
0.8 &	0.01017 &	0\% &	0\% \\
0.85 &	0.0091 &	0\% &	0\% \\
0.9 &	0.0094 &	0\% &	0\% \\
0.95 &	0.0095 &	0\% &	0\% \\
 \bottomrule
\end{tabular}
\label{tab:alpha}
\end{table}

\begin{table}[ht]
\centering
\caption{\textbf{Impact of the split ratio of training/calibration data on the tokamak fusion reactor.}}
\begin{tabular}{@{}l|c|cc@{}}
\toprule
$\alpha$    & $\J$ $\downarrow$   & $\R_{\text{sample}}$ $\downarrow$ & $\R_{\text{time}}$ $\downarrow$ \\ \midrule
0.005 &	0.0124 &	4\% &	0.03\% \\
0.01 &	0.0093 &	0\% &	0\% \\
0.02 &	0.0094 &	0\% &	0\% \\
 \bottomrule
\end{tabular}
\label{tab:split_ratio}
\end{table}

\begin{table}[ht]
\centering
\caption{\textbf{Impact of the split ratio of the weight of objective $\gamma$ on the 2D incompressible fluid.}}
\begin{tabular}{@{}l|c|cc@{}}
\toprule
$\gamma$    & $\J$ $\downarrow$ & $SVM$ $\downarrow$ & $\R$ $\downarrow$ \\ \midrule
0.01 &	0.3548 &	0 &	0\% \\
0.1 &	0.4953 &	0.004 &	2\% \\
0.3 &	0.498 &	0.003 &	2\% \\ \bottomrule
\end{tabular}
\label{tab:gamma}
\end{table}

\section{Visualization of Experiment Results}
\subsection{1D Burgers' Equation}
\begin{figure}[H]
    \centering
    \begin{subfigure}{0.7\textwidth}
        \centering
        \includegraphics[width=0.77\textwidth]{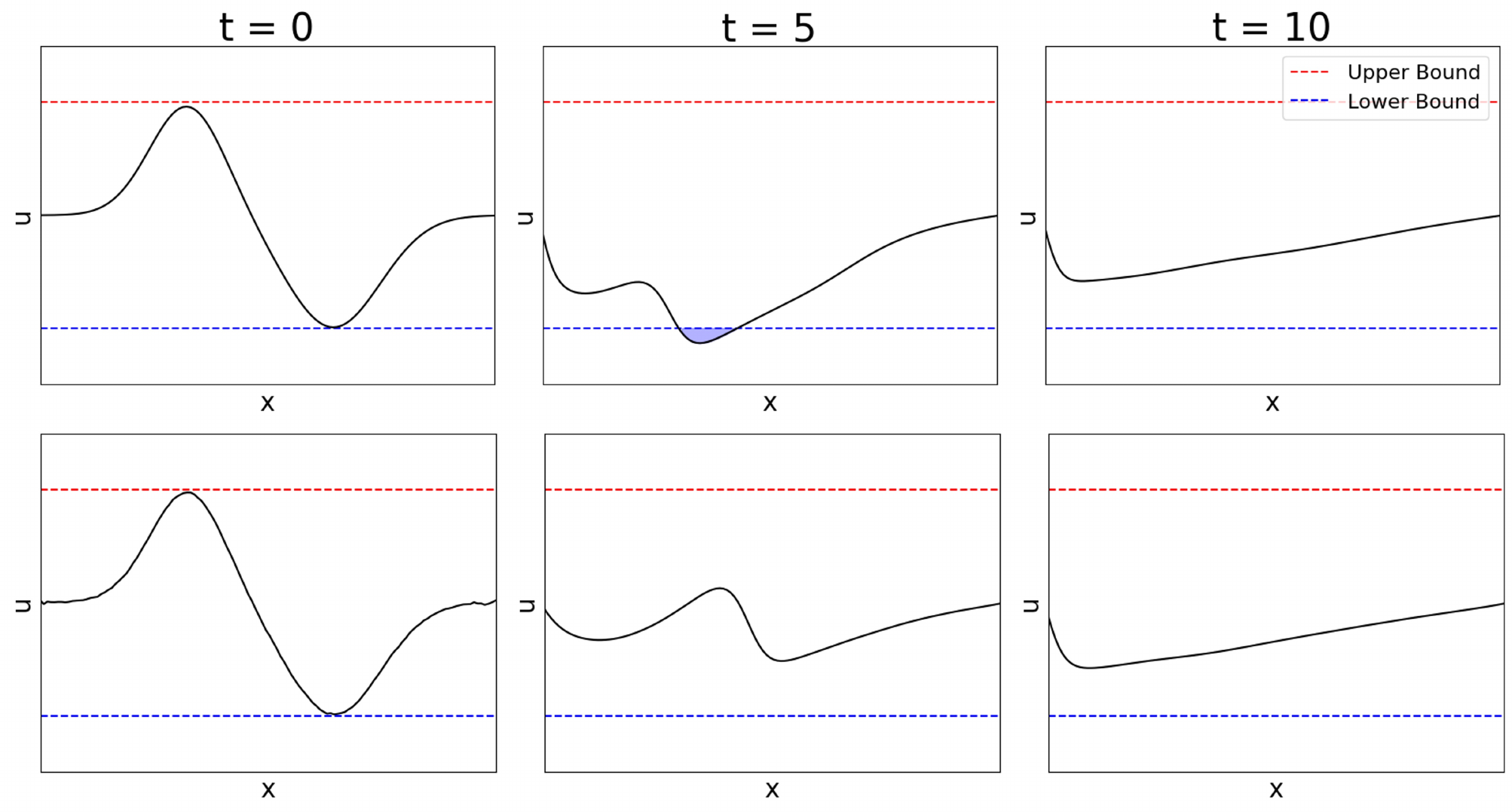}
    \end{subfigure}
    % \vspace{2pt}
    
    \begin{subfigure}{0.7\textwidth}
        \centering
        \includegraphics[width=0.77\textwidth]{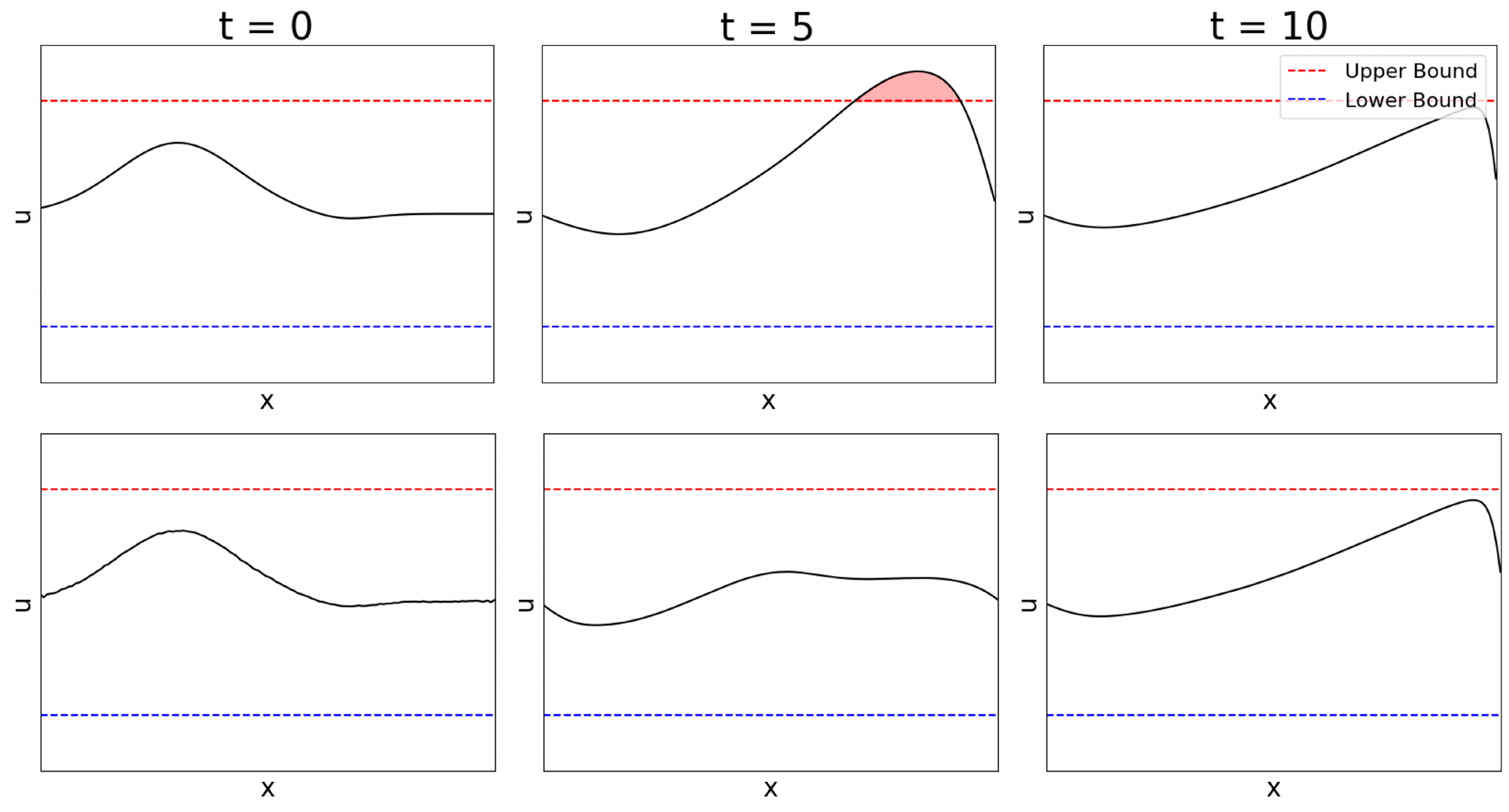}
    \end{subfigure}
    % \vspace{2pt}
    
    \begin{subfigure}{0.7\textwidth}
        \centering
        \includegraphics[width=0.77\textwidth]{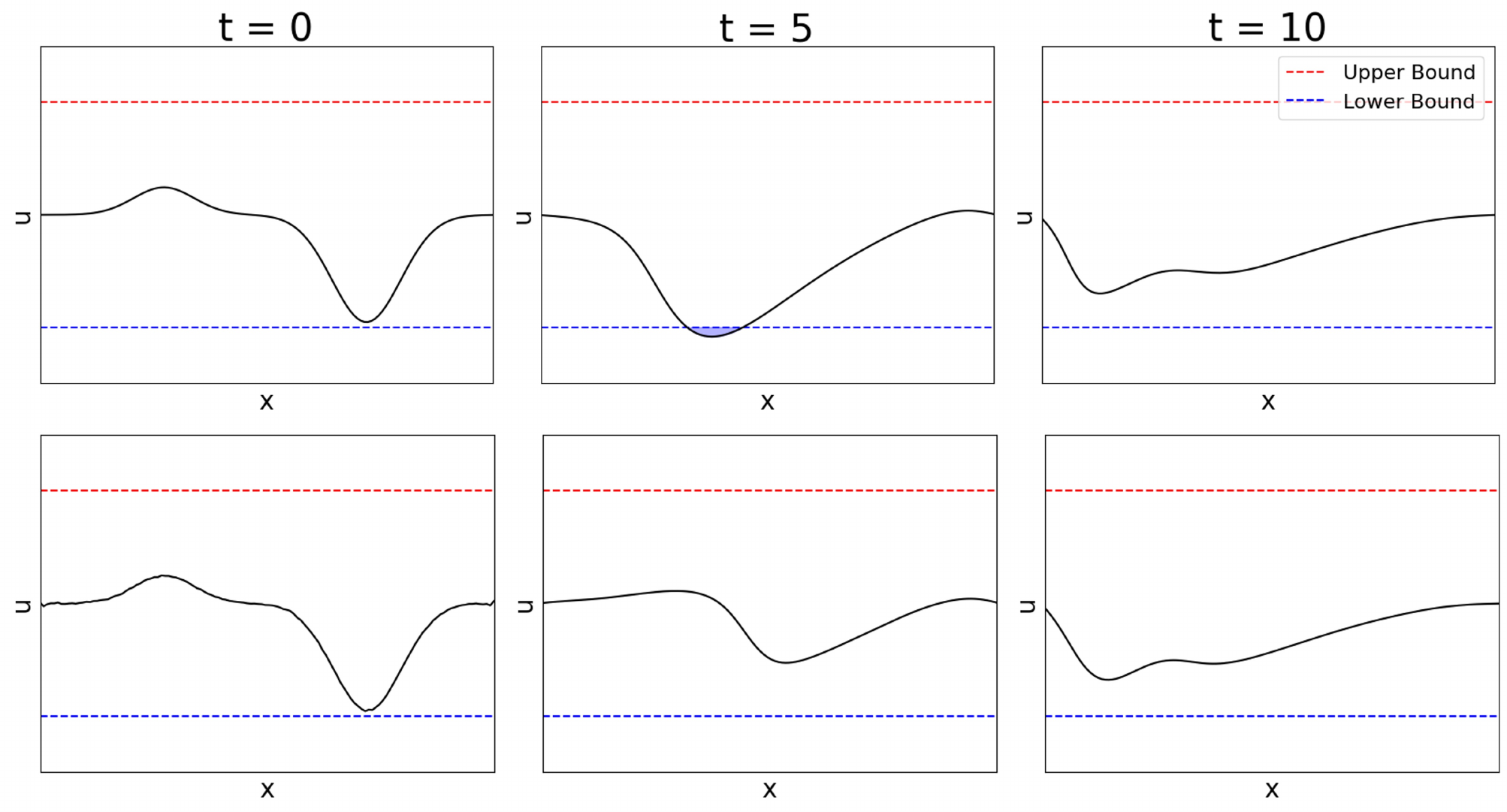}
    \end{subfigure}
    % \vspace{2pt}
    
    \caption{\textbf{Visualization of the 1D Burgers’ equation.}}
    \label{fig:1d_add}
\end{figure}

In this section, we provide additional visualizations of the control results for the 1D Burgers' equation, as shown in Figure \ref{fig:1d_add}. In these figures, the top row represents the original trajectories corresponding to the control targets, while the bottom row displays the trajectories controlled by \proj. It can be observed that \proj successfully controls the trajectories, preventing boundary violations and guiding them to the desired final state.

% \begin{figure}[ht]
%     \centering

%    % \begin{subfigure}{\textwidth}
%    %      \includegraphics[width=1\textwidth]{fig/1d_example_13.pdf}
%    %      }
%    %  \end{subfigure}
%     \subfigure{
%         \includegraphics[width=1\textwidth]{fig/1d_example_13.pdf}
%         }

%     \subfigure{
%         \includegraphics[width=1\textwidth]{fig/1d_example_17.pdf}
%         }
    
%     \subfigure{
%         \includegraphics[width=1\textwidth]{fig/1d_example_23.pdf}
%         }
    
%     \caption{\textbf{Visualization of the 1D Burgers’ equation.}}
%     \label{fig:1d_add}
% \end{figure}

\subsection{2D Incompressible Fluid}
Here, we provide additional visualizations of the control problems of 2D incompressible fluid. From the figures, we can observe that \proj can successfully control the smoke to avoid the red hazardous region and reach the target bucket as well.

\begin{figure}[H]
\begin{center}
    \includegraphics[scale=0.4]{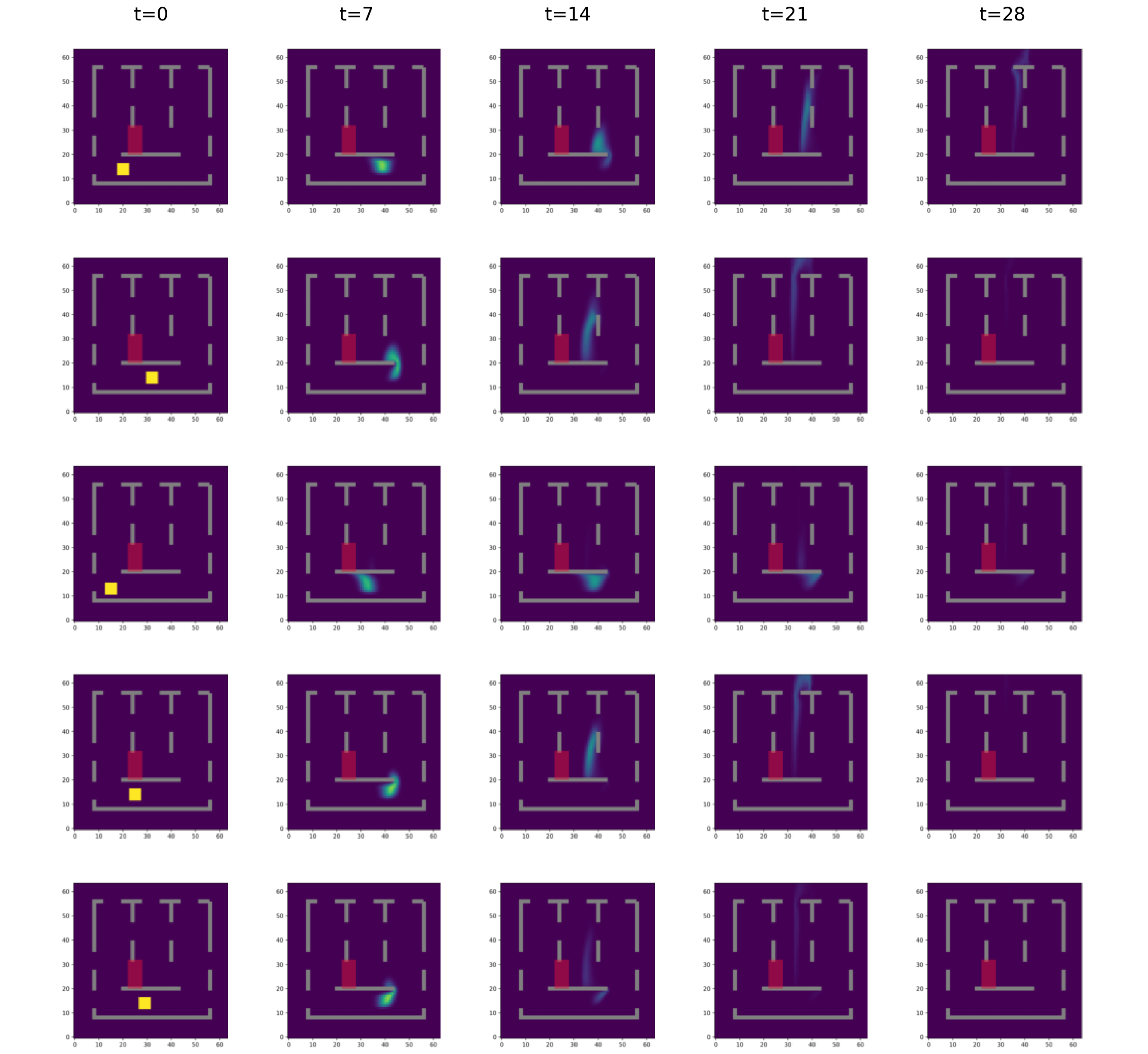}
\end{center}
\caption{\textbf{Visualizations of the 2D incompressible fluid control problem.}.}
\label{fig:2d_add}
\end{figure}

\section{Additional Details for Conformal Prediction}
\label{app:cp}
Conformal prediction is a flexible framework that provides prediction intervals with guaranteed coverage probabilities for new, unseen data points, under the assumption that the data are exchangeable.

\textbf{Theoretical Foundations} The exchangeability assumption is a cornerstone of conformal prediction. It requires that the order of the data points does not affect their joint distribution, meaning that any permutation of the indices yields an identical distribution. In particular, exchangeability holds for independent and identically distributed (i.i.d.) samples, a common assumption in machine learning tasks. Ensuring exchangeability guarantees the validity of the prediction intervals constructed using conformal prediction.

\textbf{Implementation Details} To implement conformal prediction, the dataset is first split into two subsets: a proper training set ($D_\text{train}$) and a calibration set ($D_\text{cal}$). A predictive model \({\mu}_\theta\) is trained on the training set using a specified learning algorithm \(\mathcal{A}\). Once trained, the model generates predictions for the calibration set. These predictions are used to compute `conformity scores', which measure the model's accuracy for each calibration point. Specifically, for each instance \(i\) in the calibration set, the conformity score \(S_i\) is defined as:
   \[
   S_i = |{\mu}_\theta(X_i) - Y_i|, \quad i \in D_\text{cal}.
   \]
Additionally, a worst-case score of \(\infty\) is included to account for extreme scenarios. 

Then \(1 - \alpha\) quantile \(q_{1-\alpha}(S)\) of the set of conformity scores is calculated, where \(\alpha\) represents the desired significance level (\emph{e.g.}, $\alpha=0.05$ for 95\% confidence interval).

Given a new data point \(X_{n+1}\), the prediction interval for its corresponding output is calculated as:
   \[
   \hat{C}_\alpha(X_{n+1}) = \left[{\mu}_\theta(X_{n+1}) - q_{1-\alpha}(S), {\mu}_\theta(X_{n+1}) + q_{1-\alpha}(S)\right].
   \]
This interval provides an estimate for the range within which the true value \(Y_{n+1}\) is expected to lie, with a coverage probability of at least \(1 - \alpha\). Thus, conformal prediction offers a flexible and robust method for constructing prediction intervals that account for both the model’s accuracy and the variability in the data.

\textbf{Theoretical Guarantees} Conformal prediction provides theoretical guarantees for finite samples \citep{vovk2005algorithmic, Lei2016DistributionFreePI}. Specifically, for any new data point, the prediction interval satisfies the following probabilistic bound:
   \[
   P(Y_{n+1} \in \hat{C}_\alpha(X_{n+1})) \geq 1 - \alpha.
   \]
This ensures that the true label $Y_{n+1}$ will fall within the predicted interval at least $1 - \alpha$ percent of the time. This framework, based on the assumption of exchangeability, provides a robust method for generating reliable prediction intervals, even in settings with limited sample sizes.

\section{Details of the Method Implementation}

\subsection{Conditionally Sample $\pu(\w)$}
In our proposed algorithm, we need to sample from the conditional distribution \( p(\u|\w) \) with the model that learns the joint distribution \( p(\u, \w) \). To achieve this, at each denoising step of the sampling process, we replace the noisy \(\w\) in the input of the denoising network with the actual clean \(\w\) that serves as the condition \citep{chung2023diffusion}. In fact, this situation represents a special case of the data distribution that the denoising network encounters during training, where the \(\u\) part is noisy, while the \(\w\) part remains noise-free.

\subsection{Algorithm of Post-training}
\label{app:algpost}
Here we provide the specific algorithm of post-training in Algorithm \ref{algpost}.

\begin{algorithm}[b]
    \small
    % \scriptsize
    \caption{Post-training of \proj}
    \label{algpost}
    \begin{algorithmic}[1]
    %  \SetAlgoLined
    \STATE \textbf{Require} Training set $D_{\text{train}}$, Calibration set $D_{\text{cal}}$, coverage probability $\alpha$, number of epochs $N$, number of updating steps per epoch $M$ \\
    \FOR{$n = 1, \ldots, N$} 
        \STATE Compute the shifted score set $\Tilde{\S}$ with $D_{\text{cal}}$ \small{\color{gray}// Eq. \ref{eq:weightedscore}} \\
        \STATE Get the uncertainty quantile $Q(1-\alpha;\Tilde{\S})$ \\
        \FOR{$m = 1, \ldots, M$} 
            \STATE Take gradient descent step on $\nabla_{\theta}\mathcal{L}_{\text{post-train}}$ with $D_\textrm{train}$ \small{\color{gray}// Eq. \ref{eq:posttrain_obj}} 
        \ENDFOR
\hspace{0.8cm} \\
    \ENDFOR \\
    \STATE \textbf{return} $\theta$
    \end{algorithmic}
\end{algorithm}
    
\subsection{Algorithm of Inference-time Fine-tuning}
\label{app:alg}
In this subsection, we provide the entire algorithm of inference-time fine-tuning in Algorithm \ref{alg}.

%%%%%%%%%%%%%%%%%
\begin{algorithm}[ht]
    \small
    % \scriptsize
    \caption{Inference of \proj}
    \label{alg}
    \begin{algorithmic}[1]
    %  \SetAlgoLined
    \STATE \textbf{Require} Calibration set $D_{\text{cal}}$, coverage probability $\alpha$, number of iterations $N$ \\
    \FOR{$n = 1, \ldots, N$} 
        % \STATE Compute score set $\S$ with $D_{\text{cal}}$ \small{\color{gray}// Eq. \ref{eq:scoreset}} \\
        % \STATE Compute normalized weights $\hat{\omega}$ with $D_{\text{cal}}$ \small{\color{gray}// Eq. \ref{eq:nweight}} \\
        \STATE Compute the shifted score set $\Tilde{\S}$ with $D_{\text{cal}}$ \small{\color{gray}// Eq. \ref{eq:weightedscore}} \\
        \STATE Get the uncertainty quantile $Q(1-\alpha;\Tilde{\S})$ \\
        \STATE Sample the control sequence $\w$ with guidance $\G$ \small{\color{gray}// Eq. \ref{eq:guidance}} \\
        % \STATE Sample $\pu(\w)$ conditionally to get $\score(\pu(\w))$ \\
        % \STATE Compute $\score_+(\pu(\w))$ with conditionally sampled $\pu(\w)$ \small{\color{gray}// Eq. \ref{eq:c_upper}} \\
        \STATE Take gradient descent step on $\nabla_{\theta}\mathcal{L}_{\text{fine-tune}}$ \small{\color{gray}// Eq. \ref{eq:finetune}} 
\hspace{0.8cm} \\
    \ENDFOR \\
    \STATE Sample the control sequence $\w$ with guidance $\G$ \small{\color{gray}// Eq. \ref{eq:guidance}} \\
    \STATE \textbf{return} $\w$
    \end{algorithmic}
\end{algorithm}

\section{Additional Details for 1D Burgers' Experiment}
\subsection{Experiment Setting}\label{app:1dexp}
Following the previous works \citep{holl2020learning, wei2024generative}, we generate the 1D Burgers' equation dataset. This equation is formulated as follows:
\begin{eqnarray}
% \[
\begin{cases}
\label{eq:burgers}
    \frac{\partial \u(t,x)}{\partial t} = -\u(t,x)\cdot \frac{\partial \u(t,x)}{\partial x}+\nu\frac{\partial^2 \u(t,x)}{\partial x^2} + \w(t,x)  &\text{in } [0,T] \times \Omega \\
    \u(t,x) =0  \quad\quad\quad\quad\quad\quad\quad\quad &\text{on } [0,T] \times \partial\Omega    \\
    \u(0,x) = \u_0(x) \quad\quad\quad\quad\quad\quad &\text{in } \{t=0\} \times \Omega,
\end{cases}
% \]
\end{eqnarray}
where $\nu$ denotes the viscosity parameter, while $\u_0$ signifies the initial condition. We set $\nu=0.01$, $T=1$ and $\Omega=[0,1]$. 

During inference, alongside the control sequence $\w(t,x)$, our diffusion model generates states $\u(t,x)$. Our reported evaluation metric $\J$ is always computed by feeding the control $\w(t,x)$ into the ground truth numerical solver to get $\u_{\text{g.t.}}(t,x)$ and computed following Eq. (\ref{eq:burgers_obj_J_actual}). Following Eq. (\ref{eq:burgers_safety_score}), we consider the safety constraint and define the safety score as $s$. In our experiment, the bound of safety score $s_0$ is set to 0.64, $89.7\%$ of samples are unsafe among the training set, $90\%$ of samples are unsafe among the calibration set and all of the samples in the test set are unsafe. The details of the 1D Burgers' equation dataset for safe PDE control problem are listed in Table \ref{tab:1d_data}.

\begin{table}[ht]
\centering
\caption{\textbf{Details of 1D Burgers' equation dataset.}}
\begin{tabular}{@{}cccc@{}}
\toprule
                    & Training Set & Calibration Set & Test Set \\ \midrule
Unsafe Trajectories & 34,985       & 900             & 50       \\
Safe Trajectories   & 4,015        & 100             & 0        \\ \bottomrule
\end{tabular}
\label{tab:1d_data}
\end{table}

\subsection{Model}
The model architecture in this experiment follows the Denoising Diffusion Probabilistic Model (DDPM) \citep{ho2020denoising}. For control tasks, we condition on \(u_0\), \(u_T\) and apply guidance to generate the full trajectories of \(u_{[0,T]}\), \(f_{[0,T]}\) and the safety score \(s\). The hyperparameters for the 2D-Unet architecture are recorded in Table \ref{tab:1d_hyperparameters}.

\begin{table}[ht]
  \begin{center}
    \caption{\textbf{Hyperparameters of 2D-Unet architecture in 1D experiment.}}
     \label{tab:1d_hyperparameters}
    \begin{tabular}{l|l}
    \multicolumn{2}{l}{}\\
    \hline
      \text {Hyperparameter Name} & {Value}\\ \hline
        Initial dimension          & 128           \\
        Convolution kernel size    & 3             \\
        Dimension multiplier       & {[}1,2,4,8{]} \\
        Resnet block groups        & 1             \\
        Attention hidden dimension & 32            \\
        Attention heads            & 4             \\
        Number of training steps   & 200000        \\
        DDIM sampling iterations   & 100           \\
        $\eta$ of DDIM sampling    & 1             \\ \hline
   \end{tabular}
  \end{center}
\end{table}

\section{Additional Details for 2D Fluid Experiment}
\subsection{Experiment Setting}
\label{app:2dexp}
The 2D environment is modeled by the Navier-Stokes equation:
\begin{eqnarray}
\begin{cases}
&\frac{\partial \mathbf{v}}{\partial t} + \mathbf{v} \cdot \nabla \mathbf{v} - \nu \nabla^2 \mathbf{v} + \nabla p = f, \\
&\nabla \cdot \mathbf{v} = 0, \\
&\mathbf{v}(0, \mathbf{x}) = \mathbf{v}_0(\mathbf{x}),
\end{cases}
\end{eqnarray}
where $\mathbf{v}$ is the velocity, $p$ is the pressure, $f$ is the external force and $\nu$ is the viscosity coefficient. 

Following works from \citet{holl2020learning, wei2024generative, hu2024wavelet}, we use the package \texttt{PhiFlow} to generate the 2D incompressible fluid dataset. The control objective and data generation is the same as before \citep{wei2024generative}. The main difference between our data and previous ones is that we consider the safety constraint here. We define the safety score as the percentage of smoke passing through a specific region. This reflects the need to limit the amount of pollutants passing through certain areas in real-world scenarios, such as in a watershed.

We simulate the fluid on a 128$\times$128 grid. The selected hazardous region is \([44, 36] \times [40, 64]\). Since the optimal path for smoke, starting from a left-biased position, is likely to pass through this hazardous region, this poses a greater challenge for the algorithm: how to balance safety and achieving a more optimal objective, making this a more difficult problem.

\subsection{Model}
In this paper, the design of the three-dimensional U-net we use is based on the previous work \citep{ho2022video}. In our experiment, we utilize spatio-temporal 3D convolutions. The U-net consists of three key components: a downsampling encoder, a central module, and an upsampling decoder.

The diffusion model conditions on the initial density and uses guidance as previous mentioned to to generate the full trajectories of density, velocity, control, the objective $\J$ and the safety score. As $\J$ and $\score$ are scalers, we repeat them to match other channels. The hyperparameters for the 3D U-net architecture are listed in Table \ref{tab:3d-Unet}.

\begin{table}[ht]
  \begin{center}
    \caption{\textbf{Hyperparameters of 3D-Unet architecture in 2D experiments}.}
     \label{tab:3d-Unet}
    \begin{tabular}{l|l} % <-- Alignments: 1st column left, 2nd middle and 3rd right, with vertical lines in between
    \multicolumn{2}{l}{}\\
    \hline
      \text {Hyperparameter Name} & {Value}\\
      \hline
      % Input shape & $\left[batch\ size,frames,channels,64,64\right]$ \\
      % Output shape & $\left[batch\ size,frames,channels,64,64\right]$\\
      Number of attention heads & 4 \\
      Kernel size of conv3d & (3, 3, 3)   \\
      Padding of conv3d & (1,1,1)  \\
      Stride of conv3d & (1,1,1)  \\
      Kernel size of downsampling & (1, 4, 4)   \\
      Padding of downsampling & (1, 2, 2)  \\
      Stride of downsampling &  (0, 1, 1)  \\
      Kernel size of upsampling & (1, 4, 4)   \\
      Padding of upsampling & (1, 2, 2)  \\
      Stride of upsampling &  (0, 1, 1)  \\
      Number of training steps &  200000  \\
      DDIM sampling iterations &  100  \\
      $\eta$ of DDIM Sampling & 1 \\
      Intensity of guidance in control & 100 \\
      Weight of safety term in guidance & 10000 \\
      \hline
   \end{tabular}
  \end{center}
\end{table}

\section{Additional Details for Tokamak Fusion Reactor}
\subsection{Experiment Setting}
Following the previous work \citep{seo2022development}, the environment used in this experiment is a data-driven simulator designed to replicate the plasma behavior in the KSTAR tokamak. It is constructed using long short-term memory (LSTM) neural networks to capture the time-dependent dynamics of plasma, such as transport processes and flux diffusion. The LSTM-based model is trained on five years of experimental data from KSTAR, incorporating various tokamak control variables as inputs, including plasma current, toroidal magnetic field, line-averaged density, neutral beam injection and electron cyclotron heating powers, and boundary shape parameters.

The simulator predicts the evolution of key zero-dimensional (0D) plasma parameters ($1-\alpha_p$, $l_i$, and $q_{95}$) on a 100 ms timescale in an autoregressive manner, using the current state and control inputs. This enables fast and experimentally relevant predictions of plasma responses. Using this model avoids the high costs and risks associated with real tokamak experiments while ensuring sufficient accuracy and efficiency for training.

% \subsection{Model}

\section{Baselines}

\subsection{CDT}
Constraints Decision Transformer (CDT) \citep{liu2023constrained} models control as a multi-task regression problem, extending the Decision Transformer (DT) \citep{chen2021decision}. It sequentially predicts returns-to-go, costs-to-go, observations, and actions, making actions dependent on previous returns and costs. The authors propose two techniques to adapt the model for safety-constrained scenarios:

\begin{enumerate}
\item \textbf{Stochastic Policy with Entropy Regularization}: This technique aims to reduce the risk of constraint violations due to out-of-distribution actions. In a deterministic policy, the model selects a single action based on its learned policy, which may result in unsafe actions when faced with states not well represented in the training data. By using a stochastic policy, the model samples actions from a distribution, encouraging the exploration of a wider action space. Entropy regularization further enforces diversity in the sampled actions, making the model more robust in uncertain or underrepresented situations. This approach reduces the likelihood of selecting unsafe actions when faced with states outside the distribution of the training set.

\item \textbf{Pareto-Frontier-Based Data Augmentation}: The technique tries to resolve the conflict between maximizing returns and adhering to safety constraints by leveraging a Pareto-frontier of the training data. The Pareto-frontier consists of trajectories that provide the highest possible return under specific safety constraints. Fitting a polynomial to the Pareto-frontier helps identify conflicting high-return and safety constraint pairs, which are then used for augmentation. The augmentation generates synthetic trajectories by relabeling safe trajectories from the Pareto-frontier with higher returns and assigning higher or equal safety constraints. This encourages the model to imitate the most rewarding, safe trajectories when the desired return given the safety constraint is infeasible.

\end{enumerate}

In the 2D incompressible fluid experiment, the model fails to extrapolate safe trajectories with higher returns due to the training data \textbf{covering a broad range of costs, while the desired cost lies in a narrow range}. The augmentation treats all safe trajectories on the Pareto-frontier equally, without emphasizing the region of interest. 
% To investigate, we filtered the training data to include only safe trajectories under the desired safety bound and retrained the model, showing that the data complexity exceeds CDT's capacity. 
This complexity, however, could potentially be addressed by \proj, which is indicated in Table \ref{tab:2d}. We use the official CDT implementation and follow DT guidelines to sweep desired returns and cost constraints in testing time. In the 1D Burgers experiment, we modify the control objective from the original mean squared error $\J$ between the prediction and target, to an exponential form $\exp\left( -\J \right)$. This new objective is bounded within $[0,1]$, which better aligns with the reward-maximizing setup in reinforcement learning used by CDT. For the 2D incompressible fluid setup, the state, action, and cost prediction heads each consist of 3-layer MLPs with the transformer's hidden dimension as the inner size.

\begin{table}[ht]
  \begin{center}
    \caption{\textbf{Hyperparameters of 1D CDT}.}
     \label{tab:1d_cdt}
    \begin{tabular}{l|c|c}
    \hline
    {1D Burgers'} & {All data} & {Safe filtered} \\
    \hline
    State Dimension                          & 256                & 256                \\
    Action Dimension                         & 128                & 128                \\
    Hidden Dimension                         & 1024               & 1024               \\
    Number of Transformer Blocks             & 2                  & 2                  \\
    Number of Attention Heads                & 8                  & 8                  \\
    Horizon (Sequence Length)                & 5                  & 5                  \\
    Learning Rate                            & 1e-4               & 1e-4               \\
    Batch Size                               & 64                 & 64                 \\
    Weight Decay                             & 1e-5               & 1e-5               \\
    Learning Steps                           & 1,000,000          & 1,000,000          \\
    Learning Rate Warmup Steps               & 500                & 500                \\
    Pareto-Frontier Fitted Polynomial Degree & 0                  & 4                  \\
    Augmentation Data Percentage             & 0.3                & 0.3                \\
    Max Augment Reward                       & 10.0               & 10.0               \\
    Min Augment Reward                       & 1.0                & 1.0                \\
    Target Entropy                           & -128               & -128               \\
    Testing Time Sweep Returns               & 9.0, 9.9           & 9.0, 9.9           \\
    Testing Time Sweep Costs                 & 0.0, 1.0, 2.0, 3.0 & 0.0, 1.0, 2.0, 3.0 \\ \hline
    \end{tabular}
  \end{center}
\end{table}

\begin{table}[ht]
  \begin{center}
    \caption{\textbf{Hyperparameters of 2D CDT}.}
     \label{tab:2d_cdt}
    \begin{tabular}{l|c|c}
    \hline
    {2D incompressible fluid} & {All data} & {Safe filtered} \\
    \hline
    State Dimension                          & 3$\times$64$\times$64    & 3$\times$64$\times$64    \\
    Action Dimension                         & 2$\times$64$\times$64    & 2$\times$64$\times$64    \\
    Hidden Dimension                         & 512                      & 512                      \\
    Number of Transformer Blocks             & 3                        & 3                        \\
    Number of Attention Heads                & 8                        & 8                        \\
    Horizon (Sequence Length)                & 10                       & 10                       \\
    Learning Rate                            & 1e-4                     & 1e-4                     \\
    Batch Size                               & 8                        & 8                        \\
    Weight Decay                             & 1e-5                     & 1e-5                     \\
    Learning Steps                           & 1,000,000                & 1,000,000                \\
    Learning Rate Warmup Steps               & 500                      & 500                      \\
    Pareto-Frontier Fitted Polynomial Degree & 4                        & 4                        \\
    Augmentation Data Percentage             & 0.3                      & 0.3                      \\
    Max Augment Reward                       & 32.0                     & 32.0                     \\
    Min Augment Reward                       & 1.0                      & 1.0                      \\
    Target Entropy                           & -(2$\times$64$\times$64) & -(2$\times$64$\times$64) \\
    Testing Time Sweep Returns               & 18.0, 32.0               & 18.0, 32.0               \\
    Testing Time Sweep Costs                 & 0.0, 0.1, 0.2            & 0.0, 0.1, 0.2            \\ \hline
    \end{tabular}
  \end{center}
\end{table}

\subsection{BC-All}
The Behavior Cloning (BC) algorithm, introduced by \citep{pomerleau1988alvinn}, is a foundational technique in imitation learning. BC is designed to derive policies directly from expert demonstrations, utilizing supervised learning to associate states with corresponding actions. This method eliminates the necessity for exploratory steps commonly required in reinforcement learning by replicating the actions observed in expert demonstrations. One of the significant advantages of BC is that it does not involve interacting with the environment during the training phase, which streamlines the learning process and diminishes the demand for computational resources.

In this approach, a policy network is trained using standard supervised learning strategies aimed at reducing the discrepancy between the actions predicted by the model and those performed by the expert in the dataset. The commonly used loss function for this purpose is the mean squared error between the predicted actions and expert actions. The dataset for training comprises state-action pairs harvested from these expert demonstrations. In the work, we employ the implementation as \cite{liu2023datasets}.

\subsection{BC-Safe}
Following the baseline in \cite{liu2023datasets}, the BC-Safe is fed with only safe trajectories filtered from the training dataset, satisfies most safety requirements, although with conservative performance and lower rewards. Others are same as BC-All, except for the safe trajectories.

\subsection{SL-Lag}

\citet{hwang2022solving} introduces a supervised learning (SL) based method to control PDE systems. It first trains a neural surrogate model to capture the PDE dynamics, which includes a VAE to compress PDE states and controls into the latent space and another model to learn the PDE's time evolution in the latent space. To obtain the optimal control sequence, SL can compute the gradient $\nabla_{\mathbf{w}} \mathcal{J}$, where $\mathcal{J}$ is the control objective and $f$ is the input control sequence. Then iterative gradient optimization can be executed to improve the control sequence. 

To ensure that the optimal control is compatible with the hard constraint in our experiments, we follow \citet{chow2018risk} to apply the Lagrange optimization method to the constrained optimization. Specifically, we iteratively solve the optimization problem below:
\begin{align}
    \max_{\lambda \ge 0} \min_{\mathbf{w}} \mathcal{J}(\mathbf{w}) + \lambda (s(\mathbf{u}(\mathbf{w})) - c).
\end{align}

We denote the modified SL method SL-Lag. 
% We tune its hyperparameters as follows: initialized constant $f$ [0.0001, 0.001, 0.01]; learning rate [1e-3, 1e-2, 1e-1]; Lagrange multiplier iteration []; Lagrange multiplier learning rate []

% However, simply adding the safety requirement as a loss term into the control objective results in a soft constraint and is not appropriate. % how does our method apply the constraint?

\begin{table}[ht]
  \begin{center}
    \caption{\textbf{Hyperparameters of network architecture and training for SL-Lag in 1D Burgers' experiment}.}
     \label{tab:sl_1d}
    \begin{tabular}{l|l} % <-- Alignments: 1st column left, 2nd middle and 3rd right, with vertical lines in between
    \multicolumn{2}{l}{}\\
    \hline
      \text {Hyperparameter name} & {Value}  \\
      \hline
      Initialization value of $\mathbf{w}$ & 0.001 \\
      Optimizer of $\mathbf{w}$ & LBFGS \\
      Learning rate of $\mathbf{w}$ & 0.1 \\
      Initialization value of the Lagrange multiplier $\lambda$ & 0 \\
      Optimizer of $\lambda$ & SGD \\
      Learning rate of $\lambda$ & 10 \\
      Iteration of $\lambda$ & 2 \\
      Loss function & MSE \\
      \hline
    \end{tabular}
  \end{center}
\end{table}
% f_init_0.001_wrec_0.001_wf_0.0_lr_0.1_lamlr_10.0_lam_iter_2_lam_clamp_0.01

\subsection{MPC-Lag}

We adopt the MPC with stochastic gradient descent and choose the planning step $K=1$. The simulation model is the same as the one in SL-Lag. Also, the way we combine MPC with the Lagrangian method is also the same as SL-Lag. The hyperparameters are reported in Table \ref{tab:mpc_1d}.

\begin{table}[ht]
  \begin{center}
    \caption{\textbf{Hyperparameters of for MPC-Lag in 1D Burgers' experiment}.}
     \label{tab:mpc_1d}
    \begin{tabular}{l|l} % <-- Alignments: 1st column left, 2nd middle and 3rd right, with vertical lines in between
    \multicolumn{2}{l}{}\\
    \hline
      \text {Hyperparameter name} & {Value}  \\
      \hline
      Initialization value of $\mathbf{w}$ & 0.01 \\
      Optimizer of $\mathbf{w}$ & LBFGS \\
      Learning rate of $\mathbf{w}$ & 0.01 \\
      Initialization value of the Lagrange multiplier $\lambda$ & 0 \\
      Optimizer of $\lambda$ & SGD \\
      Learning rate of $\lambda$ & 10 \\
      Iteration of $\lambda$ & 2 \\
      Loss function & MSE \\
      \hline
    \end{tabular}
  \end{center}
\end{table}
% f_init_0.01_wrec_0.01_wf_0.0_lr_0.01_lamlr_10.0_lam_iter_2_lam_clamp_5e-06

\subsection{TREBI}

In \citet{lin2023safe}, the diffusion model is adopted for the planning task under safety budgets. It generates trajectory under safety constraints using classifier-guidance \citet{dhariwal_diffusion_2021} by adding a safety loss to the reward guidance following Diffuser \citet{janner2022planning}. 

However, the original setting is different from our experiments. 
In our 1D Burgers' equation control, our objective is that a certain state equals the target state, which in-painting diffusion condition \citet{janner2022planning} is more appropriate for. Furthermore, as in our method and in \citep{wei2024generative}, a conditional diffusion model can be learned to tackle the objective more directly. 
In addition, TREBI follows the setting in Diffuser where the interaction with the environment is allowed which in our experiments becomes an MPC method. Note that the reported results of our method do not involve interaction with the surrogate model (though our method can easily adapted to be an MPC method). Thus, the results of TREBI in Table \ref{table:1d} and \ref{tab:2d} have an unfair advantage.

Therefore, for 1D Burgers' experiment, we conducted different experiments on TREBI including (1) planning multiple times with interaction with the surrogate model or (2) planning only once, and with target state conditioning or target state guidance. The target state guidance + planning multiple times turned out the best and is reported in Table \ref{table:1d}. 
For 2D smoke control, the target is not a state constraint but a reward, and the planning multiple-step setting is too computationally expensive. To this end, we use reward guidance with planning one single time, which is identical to the ablation study of our method in Table \ref{tab:ablation}. The hyperparameters of the 1D experiment are reported in Table \ref{tab:TREBI_1d}, and those of 2D are in Table \ref{tab:TREBI_2d}.

\begin{table}[ht]
  \begin{center}
    \caption{\textbf{Hyperparameters of network architecture and training for TREBI in 1D Burgers' experiment}.}
     \label{tab:TREBI_1d}
    \begin{tabular}{l|l} % <-- Alignments: 1st column left, 2nd middle and 3rd right, with vertical lines in between
    \multicolumn{2}{l}{}\\
    \hline
      \text {Hyperparameter name} & {Value}  \\
      \hline
      Reward guidance intensity & 50 \\
      Safety guidance intensity & 1 \\
      Cost budget & 0.64 \\
      Number of guidance steps & 10 \\
      Denoising steps & 200 \\
      Sampling algorithm & DDPM \\
      U-Net dimension & 64 \\
      U-Net dimension mltiplications & 1, 2, 4, 8 \\
      Planning horizon & 8 steps \\
      Optimizer & Adam \\
      Learning rate & 0.0002 \\
      Batch size & 16 \\
      Loss function & MSE \\
      \hline
    \end{tabular}
  \end{center}
\end{table}

\begin{table}[ht]
  \begin{center}
    \caption{\textbf{Hyperparameters of network architecture and training for TREBI in 2D Incompressible Fluid Control experiment}.}
     \label{tab:TREBI_2d}
    \begin{tabular}{l|l} % <-- Alignments: 1st column left, 2nd middle and 3rd right, with vertical lines in between
    \multicolumn{2}{l}{}\\
    \hline
      \text {Hyperparameter name} & {Value}  \\
      \hline
      Reward guidance intensity & 5000 \\
      Safety guidance intensity & 0.01 \\
      Cost budget & 0.1 \\
      Number of guidance steps & 1 \\
      Denoising steps & 20 \\
      Sampling algorithm & DDPM \\
      3D U-Net dimension & 8 \\
      3D U-Net dimension mltiplications & 1, 2, 4, 8 \\
      Planning horizon & 8 steps \\
      Optimizer & Adam \\
      Learning rate & 0.0002 \\
      Batch size & 16 \\
      Loss function & MSE \\
      \hline
    \end{tabular}
  \end{center}
\end{table}

% \footnotemark, 

% \footnotetext{This analysis is equivalent to the classifier-based guidance used in \citet{janner2022planning}. It has to introduce a binary random variable $\mathcal{O}\sim Bern(e^r)$, where $r$ is the expected return. Then classifier-based generation can be executed by setting $\mathcal{O}$ to 1. This complicates the analysis below, but the conclusions are the same.}

% \citet{du_reduce_2023}.

% resembling the classifier-based diffusion guidance \citep{dhariwal_diffusion_2021}.

\subsection{PID}
Proportional Integral Derivative (PID) control \citep{1580152} is a versatile and effective method widely employed in numerous control scenarios. For 1D control task, we mainly implement the PID baseline adapted from \cite{wei2024generative}. More detailed configurations can be found in Table \ref{tab:ANN_PID_architecture}
\begin{table}[ht]
  \begin{center}
    \caption{\textbf{Hyperparameters of network architecture and training for ANN PID}.}
     \label{tab:ANN_PID_architecture}
    \begin{tabular}{l|l} % <-- Alignments: 1st column left, 2nd middle and 3rd right, with vertical lines in between
    \multicolumn{2}{l}{}\\
    \hline
      \text {Hyperparameter name} & {Value}  \\
      \hline
      Kernel size of conv1d & 3 \\
      Padding of conv1d & 1 \\
      Stride of conv1d & 1 \\
      Activation function & Softsign \\
      Batch size & 16 \\
      Optimizer & Adam \\
      Learning rate & 0.0001 \\
      Loss function & MAE \\
      \hline
    \end{tabular}
  \end{center}
\end{table}

%%%%%%%%%%%%%%%%%%%%%%%%%%%%%%%%%%%%%%%%%%%%%%%%%%%%%%%%%%%%%%%%%%%%%%%%%%%%%%%
%%%%%%%%%%%%%%%%%%%%%%%%%%%%%%%%%%%%%%%%%%%%%%%%%%%%%%%%%%%%%%%%%%%%%%%%%%%%%%%

\end{document}